\documentclass[lettersize,journal]{IEEEtran}
\usepackage{pdflscape}

\usepackage{algorithmic}
\usepackage{soul}
\usepackage[normalem]{ulem}
\usepackage{algorithm}
\usepackage{array}
\usepackage[caption=false,font=normalsize,labelfont=sf,textfont=sf]{subfig}
\usepackage{textcomp}
\usepackage{stfloats}
\usepackage{url}
\usepackage{verbatim}
\usepackage{graphicx}
\usepackage{cite}
\usepackage{tikz}
\usepackage{pgfplots}
\usepackage{longtable}
\usepackage{colortbl}
\usepackage{forest}
\usepackage{tikz-qtree}
\usepackage{geometry}
\usepackage{xcolor}
\usepackage{amsmath}
\usepackage{amsfonts}

\usetikzlibrary{shapes,arrows,positioning,fit,backgrounds}
\usetikzlibrary{arrows.meta} 

\usepackage{soul}
\usepackage{xcolor}

\usepackage[table]{xcolor}
\usepackage{booktabs}
\usepackage{adjustbox}
\usepackage{array}
\usepackage{graphicx}
\usepackage{multirow}
\usepackage{rotating}
\usepackage{subcaption}         

\definecolor{bestaccuracy}{RGB}{255,235,235}
\definecolor{bestf1}{RGB}{235,255,235}

\newcommand{\best}[1]{\cellcolor{yellow!25}{#1}}

\usepackage{listings}
\usepackage{ragged2e}
\pgfplotsset{compat=1.17}

\raggedright
\newcolumntype{L}[1]{>{\raggedright\arraybackslash}p{#1}}
\newcolumntype{C}[1]{>{\centering\arraybackslash}p{#1}}

\rowcolors{3}{gray!10}{white}

\begin{document}
\justifying 

\title{A Survey on SAR ship classification using Deep Learning}

\author{\IEEEauthorblockN{Ch Muhammad Awais\IEEEauthorrefmark{1}\IEEEauthorrefmark{2}\IEEEauthorrefmark{3},
Marco Reggiannini\IEEEauthorrefmark{2}\IEEEauthorrefmark{3},
Davide Moroni\IEEEauthorrefmark{2}, 
 and
Emanuele Salerno\IEEEauthorrefmark{2}}\\
\IEEEauthorblockA{\IEEEauthorrefmark{1}PhD School in Computer Science,
University of Pisa, 56126 Pisa, Italy\\}
\IEEEauthorblockA{\IEEEauthorrefmark{2}Institute of Information Science and Technologies, National Research Council of Italy, 56124 Pisa, Italy\\ \IEEEauthorrefmark{3}National Biodiversity Future Center - NBFC, Palermo, Italy}
}

\markboth{Arxiv PrePrint, 2025}%
{Shell \MakeLowercase{\textit{et al.}}: A Survey on SAR ship classification using Deep Learning}


\maketitle


\begin{abstract}
Deep learning (DL) has become a central approach for ship classification using synthetic aperture radar (SAR) imagery. This survey reviews 74 representative studies selected from 187 publications, categorizing them into a taxonomy with four main dimensions: (i) DL architectures, (ii) datasets, (iii) image augmentation, and (iv) learning techniques. We analyze how approaches such as handcrafted feature integration, data augmentation, fine-tuning, and transfer learning influence classification performance, and we summarize the use of public benchmarks including OpenSARShip and FUSARShip. The survey highlights key challenges: limited data availability, class imbalance, lack of standardized metrics, and limited interpretability of DL models. Future research directions include the development of SAR-specific DL architectures, advanced augmentation and generative approaches, integration of handcrafted and deep features, interpretable DL, and stronger interdisciplinary collaboration. By addressing these challenges, DL-based SAR ship classification can achieve greater robustness, accuracy, and transparency, ultimately strengthening maritime surveillance and operational monitoring.
\end{abstract}

\begin{IEEEkeywords}
SAR ship classification, Deep learning, Synthetic Aperture Radar
\end{IEEEkeywords}

\section{Introduction}\label{sec1}

Our planet is covered by approximately 71\% water, which facilitates over 80\% of global trade. This vast commercial activity necessitates maritime surveillance, which is a complex research field due to its extensive coverage and diverse monitoring requirements. These requirements include irregular migration, piracy, fisheries management, and traffic monitoring. Maritime surveillance also plays a crucial role in promoting environmental sustainability by providing real-time data and insights to tackle pollution, reduce emissions, conserve marine ecosystems, and ensure the long-term health of our oceans.

Many common applications of maritime monitoring exploit SAR and Automatic Identification System (AIS) data. SAR data can be collected from satellites and airplanes, with the ability to acquire information about objects on the sea surface, such as ships. The benefits of SAR are that it can operate in any light and weather conditions, it can provide images of a vast area and revisit the same area repeatedly. AIS data are transmitted from ships, received and broadcasted by ground or satellite-based stations, and contain information about the vessel, such as its unique ID, position, speed, course, and other relevant details. These data are frequently transmitted for applications such as maritime traffic management, collision avoidance, search and rescue. AIS and SAR data can be combined to create datasets that enhance maritime monitoring applications. 
\begin{figure}[t]
    \centering
    \includegraphics[width=1\linewidth]{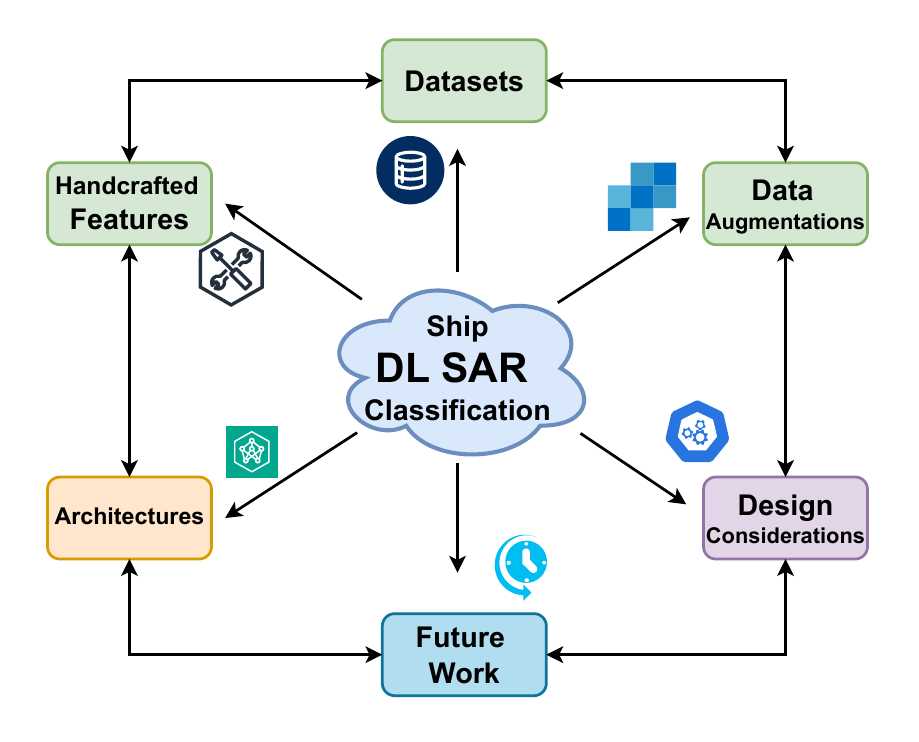}
\end{figure}

A critical feature of SAR technology is polarization.
SAR systems operate in various polarization modes, such as single-polarization (VV or HH), dual-polarization (VV and VH, or HH and HV), and fully polarized modes (VV, VH, HH, HV). Here, ``V'' stands for vertical polarization and ``H'' for horizontal polarization, with the first letter indicating the transmitted polarization and the second the received polarization. These different modes provide varying levels of detail and information about the target, influencing the type of analysis and the performance of specific DL algorithms. Different acquisition modes, such as single-look or multi-look, further influence image characteristics. Understanding these fundamental aspects of SAR imagery is essential for effective utilization of the information contained within these datasets.
Due to their broad coverage, availability, and specific benefits, SAR data can be used for several applications, such as oil spill detection \cite{espedal1999satellite, salberg2014oil, topouzelis2008oil}, ship detection \cite{li2022deep, yasir2023ship, zhang2021sar}, ship velocity estimation \cite{zilman2004speed, kang2019ship, graziano2019integration, heiselberg2023ship}, and ship classification \cite{xing2013ship, jiang2016ship, lang2015ship, wu2017novel}. In SAR images, the area represented by each pixel varies depending on the specifics of the sensing platform. For example, the Sentinel-1 C-band SAR satellite constellation provides images of Earth's surface with a spatial resolution of approximately 20 meters. This limited resolution can render vessels as only a few pixels, making object identification challenging without additional information. To address this limitation and achieve accurate target labeling, when possible, SAR-derived objects are integrated with AIS data \cite{galdelli2020integrating}.

To properly generate training sets for classification, SAR data can be matched with AIS data using several techniques \cite{chaturvedi2012ship} such as time-based matching, range-based matching and feature-based matching. 
Usually, SAR images are cropped so that each image chip represents a single detected ship, which is then labeled according to the matched AIS data. Then, different machine learning models are trained to detect and classify ships from unseen SAR images. The output is usually a set of bounding boxes in the large SAR image, indicating the presence of the detected ships and their associated categories. This approach to classification has shown good results using classical machine learning models, but DL models are also being explored for the same purpose. 

\IEEEpubidadjcol

Ships exhibit a remarkable diversity in types. For example, the FUSAR dataset \cite{hou2020fusar} encompasses 15 primary ship categories and 98 subcategories. whereas this extensive dataset proves valuable, ship classification remains a challenging task due to several factors. The limited resolution of SAR images often 
obscures the unique characteristics of different ships. Additionally, their single-channel nature, as opposed to the multichannel images commonly available from optical sensors, prevents the use of color-based image processing. Moreover, the prevalence of cargo ships, the most widely used type, introduces a significant class imbalance, making cargo ships easier to be classified compared to less represented vessels. Machine learning models often struggle with ship classification due to the absence of a clear understanding of which features are most relevant for accurate classification.

Conversely, DL approaches offer a promising solution for ship classification challenges because they can autonomously extract relevant features from satellite imagery, thus aiding vessel identification. However, the effectiveness of DL models relies heavily on crucial decisions regarding data selection, augmentation techniques, learning algorithms, and model architectures. This survey provides a comprehensive review of various techniques used to train DL models for ship classification.

Whereas existing surveys focus on SAR ship detection using DL \cite{yasir2023ship,li2022deep}, a dedicated survey on SAR ship classification using DL is currently absent. This survey addresses a critical gap within the scientific community by providing a comprehensive overview of the state of SAR ship classification using DL. It aims to bridge this gap by summarizing the existing literature, elucidating key methodologies, challenges, and advances in the field. This survey also offers insights into past developments, current trends, and future directions in SAR ship classification, thus serving as a valuable resource for researchers and practitioners in this specialized domain. The following sections outline the structure of this paper. Section 2 details the survey methodology; 
Section 3 presents our findings, resulting in a DL taxonomy; Section 4 reviews the relevant literature; Section 5 explores key insights from the literature review; Section 6 provides design considerations for DL SAR Ship classification; Section 7 discusses the key issues; Section 8 outlines possible future developments, and Section 9 concludes the paper.

\subsection*{Contributions and Scope}

\paragraph*{Contributions}
This survey gives the following contributions:
\begin{itemize}
    \item A systematic review of 187 papers, the most pertinent 74 thereof are analyzed in depth.
    These references are listed in Table~\ref{tab:paper_timeline}, and their key features are summarized in Section~\ref{sec4}.
    \item A clear taxonomy (Figure~\ref{fig:taxanomy}) covering (i) DL architectures, (ii) datasets, (iii) image augmentation, and (iv) learning techniques (including transfer learning and fine-tuning).
    \item A standardized performance table, reporting scores, classes, and datasets found in the reviewed works, with emphasis on benchmarks such as OpenSARShip and FUSARShip. The performance indices are consistently reported using F1 score and overall accuracy (OA).
    \item A synthesis of empirical trends, including the effectiveness of shallow CNNs under data scarcity, the benefits of data augmentation and fine-tuning, strategies for class imbalance (such as data splitting and tailored loss functions), and the impact of feature/polarization fusion, metric learning, multi-scale/pyramid learning, learning rate scheduling, and architecture search (NAS).
    \item Practical steps to mitigate data scarcity, including creating high-quality datasets from open-source SAR imagery, establishing AIS–SAR mapping methodologies for reliable labels, and encouraging standardization and data sharing across the community.
    \item A selection of open issues, such as data scarcity, class imbalance, standardized reporting, interpretability, and reproducibility. \item Concrete directions to guide future work in SAR ship classification.
\end{itemize}

\paragraph*{Scope}
The focus of this survey is SAR-based ship classification (category-level recognition on cropped ship chips/SAR patches) using DL. We included studies that (i) operate on SAR imagery from spaceborne or airborne platforms, (ii) consider single-, dual-, or fully-polarized data, and (iii) report F1 score and/or overall accuracy (OA), which serve as the basis for comparison in this survey\footnote{When available, F1 is treated as macro-F1; if a source reports a different F1 variant, it is indicated in the performance table. Likewise, any form of accuracy reported in the source (e.g., top-1 accuracy, classification accuracy, recognition rate) is consistently referred to as overall accuracy (OA) in this survey.}. We include discussion of handcrafted feature baselines where they inform DL design and performance. We excluded studies limited to detection or segmentation without a classification component, optical-only methods without a SAR pathway, and non-maritime Automatic Target Recognition (ATR) unless directly comparable to ship classification.

\section{Survey Methodology}\label{sec2}
\subsection{Research Questions}
The aim of this survey is to find and highlight techniques that are being used in improving ship classification through DL. The research questions are focused on finding the concerning literature:

\begin{itemize}    
    \item[-] \textbf{RQ1}: How has the use of specific DL techniques evolved over time in the field of SAR ship classification?
    
    \item[-] \textbf{RQ2}: What are the performance characteristics (e.g., accuracy, F1) achieved by different DL techniques in SAR ship classification?
    
    \item[-] \textbf{RQ3}: What is the impact of incorporating advanced techniques (such as pyramid features, feature fusion, and generative methods) on the performance of SAR ship classification using DL?
    
    \item[-] \textbf{RQ4}: What are the commonly employed network architectures (e.g., convolutional neural network (CNN), ResNet) for SAR ship classification using DL, and how do they compare in terms of performance?
    
    \item[-] \textbf{RQ5}: What are the limitations and challenges associated with the existing DL techniques for SAR ship classification, as identified in the literature?
    
    \item[-] \textbf{RQ6}: Are there any specific datasets or preprocessing techniques commonly used in the studies on SAR ship classification using DL?
    
\end{itemize}

\noindent These research questions aim to provide insights into the utilization, performance, challenges, and future directions of DL methods for SAR ship classification. 

\subsection{Search Query}\label{sec:querystring}

Search strings are crucial tools to filter through vast amounts of information and identify relevant and accurate sources. 
The search string formulated to capture studies that are specifically part of our scope was 

\begin{lstlisting}[basicstyle=\footnotesize, breaklines=true]
(Deep learning OR Neural network) AND (Ship classification OR Vessel classification OR Ship categorization) AND (Synthetic Aperture Radar OR SAR)
\end{lstlisting}

Apparently, this string addresses the use of DL approaches in SAR ship classification, including variations of terms such as DL, neural network, ship classification, vessel classification, ship categorization, synthetic aperture radar, and SAR.


\subsection{Study Selection}

\begin{figure}[t]
\centering
\begin{tikzpicture}[node distance=3mm, every node/.style={align=center}]
\node (id)    [draw, rounded corners, text width=0.9\linewidth] {Records identified: \textit{1787} \\
Duplicates removed: \textit{912}};
\node (scr)   [draw, rounded corners, below=of id, text width=0.9\linewidth] {Titles/abstracts screened: \textit{875} \\
Excluded: \textit{585}};
\node (full)  [draw, rounded corners, below=of scr, text width=0.9\linewidth] {Full-text assessed: \textit{187} \\
Excluded with reasons: \textit{113}};
\node (incl)  [draw, rounded corners, below=of full, text width=0.9\linewidth] {Studies included in synthesis: \textbf{74}};
\draw[-stealth] (id) -- (scr);
\draw[-stealth] (scr) -- (full);
\draw[-stealth] (full) -- (incl);
\end{tikzpicture}
\caption{Screening flow.}
\label{fig:prisma}
\end{figure}

To identify a relevant set of papers for the systematic literature review, the study selection process (Figure~\ref{fig:prisma}) involved the following steps:

\subsubsection*{1. Initial Search}
An initial search was conducted using the search query reported in \ref{sec:querystring} on a range of established academic databases and search engines, including Web of Science, IEEE Xplore, Scopus, Springer, Science Direct, ResearchGate and Google Scholar. This comprehensive approach ensured a thorough examination of the existing literature on the topic.

\subsubsection*{2. Screening}
Two independent reviewers screened the search results by examining the titles and abstracts of the identified papers. Inclusion and exclusion criteria were applied at this stage to determine the relevance of each paper to our research.

\subsubsection*{3. Full-Text Assessment}
The papers that passed the initial screening were retrieved in full-text format. Two independent reviewers then assessed the full texts of these papers to determine their eligibility for inclusion in our review. Any disagreements were resolved through discussion and consensus.

\subsubsection*{4. Data Extraction}
Relevant information was systematically extracted from the included papers, including details such as author(s), publication year, objectives, methodology, evaluation metrics, and key findings.

\subsubsection*{5. Quality Assessment}
The quality of the included papers was assessed using appropriate tools or checklists. This step aimed to evaluate the methodological rigor and reliability of the selected papers.

\subsubsection*{6. Data Synthesis}
Finally, the data extracted from the included papers were synthesized and analyzed to derive meaningful insights, identify trends, and answer the research questions outlined above.

\vspace{12pt}
The study selection process was conducted meticulously to ensure the inclusion of high-quality and relevant papers that contribute to the body of knowledge on SAR ship classification using DL. The inclusion criteria applied during the selection process were inspired by \cite{succi2023meta}, which derives them from the PICOC approach (Population, Intervention, Comparison, Outcome, Context) proposed by \cite{petticrew2008systematic} to ensure a focused and comprehensive selection.

\subsection{Inclusion Criteria}
This survey includes papers focused specifically on ship classification from satellite SAR images using DL techniques, algorithms or models, and appearing in peer-reviewed journals, conference proceedings or other reputable scholarly sources. Additional criteria to be satisfied were:
\begin{enumerate}
  \item Studies reporting performance metrics, evaluation results, or quantitative measures of DL model effectiveness.
  \item Studies published in English to ensure accessibility and facilitate data extraction and analysis.
  \item Studies using publicly available datasets.
  \item Studies published after 2010.
\end{enumerate}
All the papers not complying with these requirements were excluded from any further analysis.\footnote{\textit{Threats to Validity}:
\textbf{Selection bias:} despite multi-database search, some works may be missed. 
\textbf{Metric comparability:} heterogeneous splits limit strict cross-paper comparisons. 
\textbf{Reporting variance:} inconsistent metric definitions (F1 variants) and missing details can affect synthesis. 
We mitigate these by transparent inclusion criteria, deduplication, and explicit caveats in tables.

\textit{Reproducibility Checklist}: 
We provide: (i) the performance table (tables-\ref{tab:sar_small_classcount}, \ref{tab:sar_large_classcount}); (ii) search strings (section-\ref{sec:querystring}); (iii) deduplication logic summary (table-\ref{tab:search_prisma}). Each entry records dataset, split notes, and metrics (F1/OA).}



\section{Results}\label{sec3}


\subsection{Search and Screening Results}

The papers were retrieved by executing the search query (see Section~\ref{sec:querystring}) across multiple databases, followed by selection based on the predefined relevance and inclusion criteria. Per-database counts \emph{after deduplication} are reported in Table~\ref{tab:search_prisma}. The initial screening reduced the corpus from 1,847 records to 187 by removing duplicates, off-topic items, and studies that did not meet the inclusion criteria.

%

\begin{table}[t]
\centering
\caption{Search results before/after deduplication and screening.}
\label{tab:search_prisma}
\begin{tabular}{lrrr}
\toprule
Source & Initial & After dedup & Included \\
\midrule
Web of Science & 52 & 33 & 13 \\
IEEE Xplore    & 55 & 43 & 18 \\
Scopus         & 63 & 19 & 7 \\
ScienceDirect  & 87 & 2 & 1 \\
SpringerLink   & 30 & - & - \\
Google Scholar & 1460 & 90 & 35 \\
ResearchGate\footnotemark & 100 & - & - \\
\midrule
\textbf{Total (unique)} & 1787 & 187 & \textbf{74} \\
\bottomrule
\end{tabular}
\end{table}
\footnotetext{Used primarily for discovery; entries were verified against peer-reviewed sources.}

%


During the full-text assessment stage, the content of the remaining 187 papers was examined to determine their relevance to research questions and inclusion criteria. The aim of this analysis was to identify the most pertinent studies contributing significantly to the development of DL-based SAR ship classification.

By rigorous evaluation of the full texts of these papers, it was ensured that only high-quality, relevant, and valid studies were included in the final selection. This process was essential for maintaining the integrity and reliability of our literature review and for extracting meaningful insights to address the research objectives. The full-text assessment further refined the set of papers, reducing them to a final number of 74 (see Table~\ref{tab:paper_timeline}) and allowing for the derivation of well-founded conclusions and evidence-based findings.

\begin{table}[t]
\centering
\caption{Chronological Summary of SAR Ship Classification Papers}
\label{tab:paper_timeline}
\scriptsize
\begin{tabular}{@{}p{1.2cm}p{6.8cm}@{}}
\toprule
\textbf{Year} & \textbf{Published Studies} \\
\midrule
2017 & \cite{8127014} \\
2018 & \cite{8113469}, \cite{s18092929}, \cite{s18093039} \\
2019 & \cite{s19010063}, \cite{8899277}, \cite{8897831}, \cite{10.1007/978-3-030-01054-6_2}, \cite{9172933} \\
2020 & \cite{Zhu_2020}, \cite{9199258} \\
2021 & \cite{9553192}, \cite{electronics10101169}, \cite{9553116}, \cite{9146990}, \cite{9554116}, \cite{9564861}, \cite{rs13112091}, \cite{9455304}, \cite{RELEKAR20214594}, \cite{9568903}, \cite{10027936}, \cite{9170851}, \cite{9390936}, \cite{9688498} \\
2022 & \cite{9387401}, \cite{9445223}, \cite{9893336}, \cite{9180260}, \cite{9885840}, \cite{9805965}, \cite{9512781}, \cite{9782459}, \cite{app12146866}, \cite{ZHANG2022108365}, \cite{rs14235986} \\
2023 & \cite{rs15112904}, \cite{rs15112917}, \cite{rs15082138}, \cite{10149595}, \cite{10147843}, \cite{10433638}, \cite{10283157}, \cite{10189816}, \cite{10239480}, \cite{10242360}, \cite{10246308}, \cite{10315181} \\
2024 & \cite{10497597}, \cite{10457854}, \cite{10487903}, \cite{10446741}, \cite{rs16071299}, \cite{10547709}, \cite{s24247954}, \cite{10642068}, \cite{10440349}, \cite{rs16183479}, \cite{10671463}, \cite{10412205}, \cite{10641086}, \cite{10578024}, \cite{10446152}, \cite{10701968} \\
2025 & \cite{10908205}, \cite{10890957}, \cite{11045289}, \cite{Awais_2025_WACV}, \cite{10937719}, \cite{11043629}, \cite{11018437}, \cite{10960691}, \cite{awais2025classification}, \cite{awais2025feature} \\
\bottomrule
\end{tabular}
\end{table}

After the meticulous and thorough evaluation of the full texts described above, we derived the taxonomy illustrated in Figure~\ref{fig:taxanomy}, which shows models, datasets, augmentations and learning techniques used in the literature for  DL-based SAR ship classification.

\begin{figure*}[t]
\centering
\footnotesize
\begin{tikzpicture}[
  node distance = 4mm and 6mm,
  every node/.style   ={draw, rounded corners, inner sep=2pt,
                        align=left, font=\footnotesize},
  arrow/.style={-{Stealth[length=2mm]}, thick},
]

\node (root) [fill=gray!20, text width=37mm, align=center]
  {DL based SAR-Ship\\Classification};

\node (arch) [below left =of root, xshift=-28mm,
              fill=blue!10,  text width=37mm] {%
  \textbf{Architectures}\\
  \rule{34mm}{0.4pt}\\ 
  \textit{CNN lineage}\\
    — Classic (AlexNet, VGG)\\
    — Residual (ResNet, DenseNet)\\
    — Inception\\
    — Custom CNN$^{\ast}$\\[2pt]
  \textit{Transformer / Attention}\\
    — ViT, SE, CBAM\\[2pt]
  \textit{Temporal / Hybrid}\\
    — CNN–LSTM\\[2pt]
  \textit{Generative}\\
    — GAN};

\node (data) [below =of root,
              fill=yellow!12, text width=37mm] {%
  \textbf{Datasets}\\
  \rule{34mm}{0.4pt}\\ 
  \textit{Benchmark collections}\\
    — OpenSARShip (OS)\\
    — FUSAR-Ship (FS)\\[2pt]
  \textit{Satellite scenes}\\
    — COSMO-SkyMed\\
    — TerraSAR-X\\
    — PALSAR-2\\
    — GF-3\\[2pt]
  \textit{Mixed / Private}\\
    — FS+OS,\ Author-specific};

\node (aug)  [below right=of root, xshift=12mm,
              fill=orange!12, text width=37mm] {%
  \textbf{Image Augmentation}\\
  \rule{34mm}{0.4pt}\\ 
  — Geometric (rot./flip/scale …)\\
  — Noise/Spectral (speckle,\ polar jitter)\\
  — Synthetic (CutMix,\ GAN-gen.)};

\node (learn) [below=of aug, yshift=-2mm, xshift=-12mm,
               fill=cyan!12, text width=46mm] {%
  \textbf{Learning Techniques}\\
  \rule{34mm}{0.4pt}\\ 
  — Feature / Polarisation fusion\\
  — Multiscale processing\\
  — Metric \& Siamese learning\\
  — Attention mechanisms\\
  — Domain adaptation / TL\\
  — Semi- / Self-supervised\\
  — Knowledge distillation\\
  — Sampling / Imbalance\\
  — Curriculum learning\\
  — Super-resolution\\
  — Neural arch.\ search};

\draw[arrow] (root.south west) -- (arch.north east);
\draw[arrow] (root.south)      -- (data.north);
\draw[arrow] (root.south east) -- (aug.north west);
\draw[arrow] (aug.south)       -- (learn.north);

\end{tikzpicture}

\caption{Taxonomy of DL methods for SAR ship classification. 
The taxonomy organizes prior work into four dimensions: (i) architectures, (ii) datasets, (iii) image augmentation, and (iv) learning techniques. 
$^{\ast}$ denotes CNN variants specifically designed by the respective authors.}
\label{fig:taxanomy}
\end{figure*}


\subsection{Survey Findings}

The analysis of relevant literature revealed several key insights into the use of DL techniques for ship classification from SAR images. Here is a list thereof, ordered by research questions (see Section \ref{sec2}):

\begin{enumerate}

\item[RQ1] \textbf{Evolution of DL Techniques:} The adoption of CNN models initially stemmed from their success in other image processing tasks. However, as the demand for enhanced performance intensified, researchers integrated various image processing techniques and handcrafted features to augment the effectiveness of DL models \cite{9553192, 10446741, 10487903, 10578024, 10189816, 10254064}. 

\item[RQ2] \textbf{Performance Characteristics:} Accuracy is the most commonly reported performance metric in the literature. However, additional metrics should be considered, particularly when addressing long-tailed or imbalanced datasets \cite{10701968}. Relying solely on accuracy can be misleading, as it may not adequately reflect performance across different classes, especially minority categories. To address this, we extracted the reported accuracy and F1 scores from the studies and summarized them in Tables~\ref{tab:sar_small_classcount} and~\ref{tab:sar_large_classcount}. It is important to note that these scores are not directly comparable across studies due to differences in dataset selection: subsets of test data vary, and even when the same number of classes is reported, the specific class composition may differ.

\item[RQ3] \textbf{Impact of Advanced Techniques:} Techniques like pyramid features, feature fusion, curriculum learning, and transfer learning were employed to bolster ship classification performance on SAR images \cite{9568903, 10315181, 10433638, 10283157, 9885840, 8899277, Awais_2025_WACV, 10937719, awais2025feature}. Handcrafted features and pyramid features demonstrated particularly promising results. However, the research in this area remains in its nascent stages, warranting further investigation.

\item[RQ4] \textbf{Commonly Employed Network Architectures:} Specific CNN architectures, particularly deeper models such as VGG and ResNet, proved to be promising in SAR ship classification due to their ability in extracting complex features. 
Interestingly, simpler and shallow CNN models, when combined with effective image processing such as noise reduction and image enhancement, have also achieved commendable accuracy \cite{rs14235986, s24247954}. 

\item[RQ5] \textbf{Limitations and Challenges:} SAR ship classification faces several critical challenges. The scarcity and imbalance of labeled SAR image data, coupled with limitations in SAR image processing techniques, hinder model performance. Additionally, relying solely on DL networks designed for optical imagery or choosing inappropriate SAR-specific architectures can lead to suboptimal results. Furthermore, SAR-AIS matching, while valuable, has limitations due to its inherent unreliability (e.g. lack of transmission, noisy and inconsistent messages \cite{EMMENS2021114975}). The development of robust, automated SAR-AIS matching algorithms is crucial. Finally, SAR processing models can benefit significantly from tailored loss functions \cite{s18093039, 9146990, 9512781} that effectively filter out irrelevant information and focus on the significant features of target objects. Research on these tailored loss functions is an active area of development.

\item[RQ6] \textbf{Specific Datasets and preprocessing Techniques:} Studies in the literature commonly use two publicly available datasets: FUSAR-Ship \cite{hou2020fusar} and OpenSARShip \cite{li2017opensarship}. Geometric preprocessing techniques, such as cropping, rotation, flipping and scaling, are frequently employed to improve the generalizability of ship classification models. However, the necessity for these techniques arises from the challenges associated with small and low-resolution vessel representations in the data. Super-resolution techniques offer a potential avenue for addressing these limitations \cite{9884456, rs16010018, Smith_2022, 11043629, 11018437}.

\end{enumerate}

\section{Background}\label{sec4}

Several techniques are employed for SAR ship classification using DL, as illustrated in Figure~\ref{fig:taxanomy}. This section is divided into four parts. The first part explores the various DL techniques used.
The second part examines the impact of handcrafted features on the performance of DL models. As SAR images possess unique attributes, the third part discusses how these attributes are leveraged.
Finally, the fourth part investigates the use of fine-tuning.

\subsection{DL Techniques}
As apparent from Table~\ref{tab:paper_timeline}, we found the origin of SAR-based ship classification using DL in 2017 \cite{8127014},
when four different ship types detected in six VV-polarized COSMO-SkyMed scenes were classified using a simple CNN model with two convolution layers and two fully connected layers.
The images of the detected targets were cropped to $64\times 64$ chips, four scenes were then used for training, one for testing and one for validation. The results showed that the model was able to learn the cargo ships better than the others, because of
their high prevalence in the training set.
The advancement of SAR platforms over time has resulted in images with better and better spatial resolutions. 
In \cite{8113469}, CNNs are used to treat images from TerraSAR-X, with a resolution of 3 meters per pixel. Four augmentation techniques are applied to form the training set: reflections (horizontal and vertical), rotations (-10 to 10 degrees), translation (-5 to 5 pixels) and noise addition (randomly changing pixel values to zero), achieving an F1 score of 94\% by using 
multiple-resolution  images.


  Two important aspects of a DL model are explored in \cite{s18093039}: an attention based CNN model and a novel loss function for class-imbalanced datasets, named ``weighted distance measure''. A VH-polarized dataset with Ground Range Detected (GRD) images is divided into 4 classes, cargo, tanker, tug and others. The results show that the model is capable of learning class representation, achieving a recognition accuracy of 84\%.  

In \cite{10.1007/978-3-030-01054-6_2} a convolution network for detection and classification is evaluated. The input data are drawn from the PALSAR-2 L-band SAR with 10 m resolution in HV polarization. The results indicate that the CNN-based models can detect ships, but the classification performance is affected by ships that look alike.

Ship classification can also be achieved using a sequence-based approach. In \cite{Zhu_2020} a sequence input is constructed by capturing ten consecutive images of the same target at different times and rotation degrees. The authors employ a combination of CNN and long short-term memory (LSTM) with four categories of OpenSARShip data (boat, cargo, container, and oil tanker). Their findings suggest that the model learns improved class representations when using image sequences. Similarly, the authors of \cite{s24247954} first optimize the CNN design by reducing filter redundancy and network size, and then integrate LSTM layers to capture sequential dependencies and enhance discrimination. Comparative experiments against state-of-the-art CNNs (e.g., VGG-16, ResNet50, DenseNet121) show that the hybrid network achieves competitive accuracy while significantly reducing training time and complexity. These results highlight the potential of lightweight CNN–LSTM frameworks for efficient ship classification in resource-constrained settings.

\begin{figure*}[t]
\centering
\footnotesize
\definecolor{cbBlue}{RGB}{86,180,233}
\definecolor{cbGreen}{RGB}{0,158,115}
\definecolor{cbOrange}{RGB}{230,159,0}
\definecolor{cbPurple}{RGB}{204,121,167}
\definecolor{cbYellow}{RGB}{240,228,66}
\definecolor{cbGrey}{RGB}{153,153,153}

\resizebox{\linewidth}{!}{%
\begin{tikzpicture}[
  node distance = 4mm and 7mm,
  box/.style={draw, rounded corners, inner sep=3pt, align=center, text width=43mm, thick, font=\footnotesize, fill=white},
  title/.style={draw, rounded corners, inner sep=4pt, align=center, thick, font=\bfseries\footnotesize, fill=cbGrey!20},
  arrow/.style={-{Stealth[length=2mm]}, thick}
]

\node (root) [title, text width=64mm]
{DL Techniques for SAR Ship Classification};

\node (early)   [box, fill=cbBlue!12, below left =of root, xshift=-6mm]
{\textbf{Custom CNN Architectures}\\ {\scriptsize \cite{8127014,8113469,10.1007/978-3-030-01054-6_2,s18093039}}};

\node (arch)    [box, fill=cbGreen!12, below right=of root, xshift=6mm]
{\textbf{Architecture enhancements}\\ {\scriptsize \cite{9553116,10642068,9564861,rs15112917,10315181,10283157}}};

\node (temporal)[box, fill=cbOrange!12, below=of early]
{\textbf{Temporal / Hybrid (CNN--LSTM)}\\ {\scriptsize \cite{Zhu_2020,s24247954}}};

\node (light)   [box, fill=cbPurple!12, below=of arch]
{\textbf{Lightweight / NAS}\\ {\scriptsize \cite{10908205,app12146866,rs15112904}}};

\node (metric)  [box, fill=cbYellow!25, below=of temporal]
{\textbf{Metric / Few-shot / Siamese}\\ {\scriptsize \cite{9199258,10147843,10433638,9512781}}};

\node (train)   [box, fill=cbBlue!12, below=of light]
{\textbf{Training strategies}\\ {\scriptsize \cite{9805965,10446152,10239480,Awais_2025_WACV}}};

\node (data)    [box, fill=cbGreen!12, below=of root]
{\textbf{Data-centric (SR, domain)}\\ {\scriptsize \cite{11018437,11043629,10242360}}};

\draw[arrow, cbGrey] (root.south west) -- (early.north east);
\draw[arrow, cbGrey] (root.south east) -- (arch.north west);
\draw[arrow, cbGrey] (root.south west) -- (temporal.north east);
\draw[arrow, cbGrey] (root.south east) -- (light.north west);
\draw[arrow, cbGrey] (root.south west) -- (metric.north east);
\draw[arrow, cbGrey] (root.south east) -- (train.north west);
\draw[arrow, cbGrey] (root) -- (data);


\end{tikzpicture}%
}
\caption{Overview of DL techniques applied to SAR ship classification. The methods are grouped into major categories, including core Custom CNN Architectures, architecture enhancements, temporal and hybrid approaches, lightweight and NAS-based models, metric and few-shot learning, training strategies, and data-centric methods. Representative studies for each category are indicated by citation keys.}
\label{fig:dl_techniques_map_compact_colored}
\end{figure*}

A CNN, when not fully connected, can be used as a feature extractor. The extracted features can then be fed to a classifier or further processed using methods like metric learning. Metric learning involves training a model to measure the similarity or distance between data points, often to cluster extracted features in a way that makes them distinguishable, thereby enhancing the model's performance. In \cite{9199258}, the performance of a CNN-only approach is compared to a CNN combined with metric learning. The OpenSARship dataset is used, restricted to bulk carrier, container, and oil tanker classes, focusing on data with larger sample sizes. The features extracted using the CNN are then enhanced by concatenating additional features learned using metric learning. The results show that the CNN combined with metric learning outperforms the simple CNN. 

In \cite{10147843}, a novel Hierarchical Embedding Network with Center Calibration (HENC) is proposed for few-shot ship classification. HENC uses a CNN to extract hierarchical features at different scales, achieving an accuracy of 81.03\% on a six-category OpenSARShip dataset (including cargo, tanker, tug, fishing, dredging and others) with few-shot training, thus demonstrating the effectiveness of deep models in learning class representations even with limited data. 

In \cite{10433638}, a deep metric learning approach is discussed, which leverages a pairwise representation fusion for multi-proxy learning and introduces a novel distribution loss to refine the sample distribution. This allows for both clustering similar classes together and training the model to make a clear distinction between all classes. The effectiveness of this three-pronged approach is evaluated on two optical ship datasets (FGSC-23 and FGSCR-42) and one SAR dataset (5-class FUSARShip).

In \cite{10315181}, the authors introduce HDSS-Net, a novel DL architecture for ship classification in SAR images. This addresses challenges such as large variations in ship size, the elongated shapes of ships in SAR images, and intraclass diversity alongside interclass similarity. HDSS-Net employs a hierarchical design with three key modules. The first, a feature aggregation module (FAM), creates a multi-scale feature pyramid. The second, a Feature Boost module (FBM), captures the elongated features of ships through rectangular convolutions. Finally, a spherical space classifier (SSC) processes features in a spherical space to enhance class separability. Evaluations on OpenSARShip (three and six classes) and FUSAR-Ship (seven classes) demonstrate that HDSS-Net outperforms traditional CNN methods, highlighting its potential value for maritime surveillance applications.

There exists a trade-off between spatial resolution and the cost of SAR imagery. Developing algorithms that achieve accurate results using medium-resolution data is essential for broader applications. DenseNet and metric learning, with a novel loss function incorporating deep metric learning \cite{9146990}, are used in \cite{hoffer2015deep} to perform SAR ship classification on three-class (tanker, container, bulk carrier) and five-class (tanker, container, bulk carrier, cargo, general cargo) datasets from OpenSARShip. To help the model learn accurate class representations, image triplets (an anchor image, a positive image, and a negative image) are used, and data augmentation techniques such as flipping, rotation, translation and Gaussian noise addition are applied to address overfitting. The results demonstrate superior performance in both the three-class and five-class experiments.

Each CNN layer stores important information about features, but some information may be lost when features pass from one layer to another. In \cite{9553116}, a novel multi-scale CNN (MS-CNN) is proposed to utilize all the information from each convolutional layer. Using three classes from the OpenSARship dataset (bulk carrier, container ship and tanker), the MS-CNN, integrated with information from each convolutional layer, achieves a 5\% better performance than the same CNN without integration. This improvement suggests that information loss occurs within the network. 

In \cite{rs15112917}, a novel deep network is proposed, with an inception-residual controller (IRC) module to improve the performance. The authors train the model on a five-category dataset from FUSARShip. The data undergo homogenization, normalization, and augmentation techniques such as flipping and rotation and, after 50 training epochs, the model with IRC achieves an accuracy of 98.71\%, indicating that deep models trained on pre-processed data can significantly improve the classification performance.

Lightweight CNNs have been explored as means to reduce the computational burden of ship classification from high-resolution SAR data while maintaining competitive accuracy. A compact architecture called LN-SCNet is proposed in \cite{10908205}. It integrates a preprocessing module to suppress speckle noise, dilated convolutions for initial feature extraction, and incorporates a feature fusion module that combines high-resolution features for small ships with deeper semantic features for large ships. To mitigate class imbalance, a seesaw loss is introduced, dynamically adjusting gradients across classes. Evaluated on 8 classes of the FUSAR-Ship dataset, LN-SCNet achieves 76.1\% accuracy and an F1 score of 56.7\%, outperforming other lightweight designs such as MobileNet and ShuffleNet while requiring only a fraction of their parameters and floating-point operations. These results suggest that carefully designed lightweight networks can offer an effective trade-off between efficiency and performance in SAR ship classification.

 A plug-and-play Multi-scale Feature Spatial Attention Fusion (MFSAF) module is introduced in \cite{10642068} to enhance CNN performance in SAR ship classification. This module integrates spatial attention with feature alignment, enabling CNNs to better capture ship features across multiple scales. Applied to several backbones (e.g., VGG-11, ResNet-18, DenseNet-121, EfficientNetV2, ConvNeXt), the MFSAF module consistently outperforms the baseline models, 
 achieving an 81.9\% accuracy on OpenSARShip2.0 when applied to ConvNeXt. These results highlight the adaptability of MFSAF and its effectiveness in improving spatial attention and feature fusion for robust ship classification in SAR imagery.

 A multi-scale dual-attention CNN that leverages information from convolutional layers is proposed in \cite{9564861}. The authors employ a VGG-Net architecture for feature extraction and utilize a three-class dataset (cargo, container ship, and tanker) from OpenSARShip. The methodology involves modifying the VGG network to extract features from SAR images, followed by the application of squeeze-and-excitation (SE) blocks to focus on both channel-wise and positional context. SE blocks learn the importance of feature maps, promoting informative features and suppressing less useful channels. The results demonstrate the effectiveness of this method, achieving an accuracy of 92.66\%. This success suggests the potential of attention mechanisms in this field. An enhanced vision transformer incorporating feature fusion modules to capture local information followed by SE attention is proposed in \cite{10283157}. The effectiveness of this approach is evaluated on a three-class OpenSARship dataset, achieving an accuracy of 76.25\%.

In the training stage, the convergence of a DL model is controlled by several parameters, among which the learning rate is of paramount relevance. It determines how the model adjusts its internal parameters during training, and can either be fixed throughout the process or adjusted dynamically. Three learning rate management strategies and their impact on transfer learning are explored in \cite{9805965}. The results suggest that a smooth learning rate change can improve the model's ability to understand the data, leading to enhanced classification performance. The authors also investigate the effect of freezing different layers during transfer learning on a CNN, emphasizing the importance of layer analysis before fine-tuning. The models are first trained on the ten-category MSTAR dataset (SAR images of tanks, \cite{afMSTAROverview}) and then fine-tuned on a three-category OpenSARShip dataset (cargo, container ship, and bulk carrier). The results demonstrate that incorporating learning rate scheduling and thoughtful layer freezing significantly improves the model performance. This finding underscores the importance of employing effective training strategies for optimal SAR ship classification.

Selecting optimal DL architectures for specific tasks can be challenging. Traditionally, researchers rely on exploring existing architectures and their applications. However, this approach often lacks a systematic methodology, leading to difficulties in finding the ideal architecture for a given classification problem. To address this challenge, a lightweight searched binary neural network (SBNN) is proposed in \cite{app12146866}, obtained through network architecture search (NAS), quantized for weight reduction and incorporating patch shift processing to remove noise. The authors employ three (bulk carrier, container and tanker) and six-class (bulk carrier, container ship, tanker, cargo, general cargo and fishing) datasets from OpenSARShip, containing both VV and VH polarizations. They apply data augmentation methods such as random cropping and horizontal flipping. The SBNN achieves accuracy scores of 80.03\% and 56.73\% for three and six classes, respectively. NAS not only results in superior performance but also lead to a lightweight model.  A novel mini-size searched convolutional metaformer (SCM) is introduced in \cite{rs15112904}, leveraging an updated NAS algorithm with progressive data augmentation (crop, flip, rotate, resize, cutout) to discover an efficient convolutional network baseline. This baseline is then enhanced with a transformer classifier to improve spatial awareness, and further augmented with a ConvFormer cell, combining the searched convolutional cell within a Metaformer block for superior feature extraction. Evaluated on VV- and VH-polarized datasets from OpenSARShip (3 categories, i.e.~bulk carrier, container, and tanker) and FUSARShip (4 categories, i.e.~bulk carrier, fishing, cargo, and tanker), SCM achieves accuracies of 82.06\% and 63.9\%, respectively, with a lightweight design (about half a million weights). In
\cite{9512781}, a one-shot learning approach using a siamese network is proposed. The approach employs a 7-layer, fully connected feature fusion that enables the network to classify 16 categories within the OpenSARShip dataset. This methodology involves three stages: (i) preprocessing the images and converting them to three dimensions by adding two additional channels (actual, filtered, and inverted SAR Image), (ii) creating image triplets containing an anchor image, a positive image, and a negative image, and (iii) training the model using a loss function that ensures the anchor image to be closer to the positive image and farther from the negative image. The results demonstrate the model's effectiveness for ship classification.

 The challenge of handling the long-tailed distribution problem in imbalanced SAR data is addressed in \cite{10239480}. The authors propose a novel architecture that incorporates a multi-branch expert network with a dual-environment sampling strategy. This network consists of three classifiers trained to handle different distribution scenarios: long-tailed, balanced, and inverse long-tailed. Intraclass dual-environment sampling adjusts sampling rates within each class based on classification confidence, creating varied distributions. In \cite{Awais_2025_WACV}, curriculum learning is used to address long-tailed distributions by introducing two curriculum methodologies for training, and also explores the importance of loss functions. Combined with center loss \cite{wen2016discriminative}, this method enhances robustness and accuracy, outperforming existing techniques on the MSTAR and FUSAR datasets. Experiments confirmed significant improvements in recognizing both majority and minority classes, thus demonstrating the effectiveness of the proposed approach in balancing feature learning and improving overall and average accuracy.

Overfitting remains a key limitation in SAR ship classification, particularly in few-shot and noisy data scenarios. To address this, \cite{10446152} presents a double reverse regularization strategy that leverages both offline and online self-distillation in a complementary manner. An Adaptive Weight Assignment (AWA) module is used to dynamically adjust the balance between the two sources of soft targets according to network performance, rather than relying on fixed schedules. Experiments on OpenSARShip and FUSAR-Ship show that this strategy consistently improves the performance of classical CNN backbones, achieving accuracy gains of up to 7\% over baselines and surpassing other distillation-based regularization approaches. The results suggest that adaptive distillation can provide an effective path toward improved generalization under limited training data.

Due to the resolution of SAR images, research on small-sized fishing vessels has not been given sufficient importance. However, as the resolution of SAR systems increases, there is a growing need to address the classification of fishing vessels. A novel architecture to this purpose is proposed in \cite{10242360}, consisting of three modules: (i) the multi-feature extraction (MUL) module, which extracts features using DenseBlocks and channel-wise concatenation, (ii) the feature fusion (FF) module, which aggregates features from different layers, and (iii) the parallel channel and spatial attention (PCSA) module, which applies channel and spatial attention. The paper also proposes a method for SAR and AIS integration to create a fishing vessel dataset. This methodology was tested on three fishing vessel classes, and the model achieved an accuracy of 89.7\%. 
 
 The importance of super resolution in SAR ship classification is discussed in \cite{11018437}, whereas \cite{11043629} emphasizes the importance of considering classification loss when enhancing the image resolutions of imbalanced datasets.

\subsection{Handcrafted Features}

Traditional features like histogram of oriented gradients (HOG) can be combined with features extracted by a CNN to enhance the performance of DL models that require a large amount of features. HOG features capture the distribution of edge orientations within an image region, describing its local shape and appearance. A VGG network is used in \cite{9553192} as the backbone for feature extraction and concatenate these features with HOG features on three categories (bulk carrier, container ship and tanker) from the OpenSARship dataset. The results indicate that the model that incorporates HOG features achieves 7\% better performance than the model without HOG features. 

\begin{figure}[ht]
\centering
\footnotesize
\resizebox{0.9\linewidth}{!}{%
\begin{tikzpicture}[
  node distance = 6mm and 6mm,
  every node/.style={draw, rounded corners, thick, fill=white, font=\footnotesize, align=left, text width=62mm, inner sep=3pt},
  title/.style={draw, thick, rounded corners, fill=gray!20, font=\bfseries\footnotesize, text width=70mm, align=center},
  arrow/.style={-{Stealth[length=2mm]}, thick}
]

\node (root) [title] {Handcrafted Feature Integration in SAR Ship Classification};

\node (fusion) [below=of root] {%
\textbf{Feature Fusion (Handcrafted + DL)}\\
\scriptsize HOG+CNN \cite{9553192}, HOG-ShipCLSNet \cite{9445223},\\
Attention-guided handcrafted fusion \cite{10487903},\\
Feature injection designs \cite{rs13112091},\\
Mutual learning (KSCM+FFCS) \cite{10189816}
};

\node (classical) [below=of fusion] {%
\textbf{Classical \& Signal-Domain Features}\\
\scriptsize Canny, Harris, Gabor, LBP \cite{10189816},\\
Time--frequency SST $\rightarrow$ HOG + CNN \cite{10671463}
};

\node (adv) [below=of classical] {%
\textbf{Advanced Learning Strategies}\\
\scriptsize Joint learning framework (FJL) \cite{10578024},\\
Contrastive/self-supervised fusion (DCPNet) \cite{10446741},\\
Bayesian CNNs for uncertainty \cite{10641086}
};

\node (meta) [below=of adv] {%
\textbf{Metadata \& Structural Features}\\
\scriptsize CNN+KNN with AIS length/breadth \cite{electronics10101169},\\
Part-based local+global features (MBR segmentation) \cite{10027936}
};

\draw[arrow] (root.south) -- (fusion.north);
\draw[arrow] (fusion.south) -- (classical.north);
\draw[arrow] (classical.south) -- (adv.north);
\draw[arrow] (adv.south) -- (meta.north);

\end{tikzpicture}%
}
\caption{Overview of handcrafted feature integration approaches in SAR ship classification. The methods are organized into four categories: (i) fusion of handcrafted descriptors with deep features, (ii) use of classical and signal-domain descriptors, (iii) advanced learning strategies incorporating handcrafted cues, and (iv) metadata- and structure-driven feature designs. Representative studies for each category are indicated by citation keys.}

\label{fig:handcrafted_condensed}
\end{figure}

 A method called HOG-ShipCLSNet is proposed in \cite{9445223}. This method first extracts HOG features and then learns deep features using a CNN called the multiscale classification mechanism (MS-CLS). MS-CLS extracts features from each convolutional layer. Since the features from different layers have varying dimensions, the network employs a global self-attention mechanism to focus on important features within each layer. To ensure stable training, a fully connected balance mechanism resizes the features to a uniform size and then averages them. Finally, the HOG feature fusion mechanism merges the handcrafted HOG features with the DL features before performing the classification. The authors conducted experiments on two datasets: three categories from OpenSARShip (bulk carrier, container ship, tanker) and seven categories from FUSARShip (bulk carrier, container ship, fishing, tanker, general cargo, other cargo, other). The results indicate that HOG-ShipCLSNet is effective for ship classification.   Another attention-guided neural network,  using OpenSARship, is proposed in \cite{10487903}. This network incorporates 14 handcrafted features processed by a feedforward network and combines them with features extracted by the attention-guided network. These combined features are then employed to train a classification model. The model, trained on 4 categories from OpenSARship, is evaluated on a dataset of SAR-AIS integrated Sentinel-1 SAR images. This approach achieves an accuracy of 86.3\%, demonstrating the potential effectiveness of combining handcrafted features with an attention mechanism for ship classification tasks.

Choosing effective features plays a vital role in classification performance: indeed, meaningful features help the model to be trained optimally.  Four handcrafted features (Canny, Harris, Gabor, and local binary patterns (LBP)) are compared in \cite{10189816}, where an optimal method for feature fusion is proposed. 
The method consists of two blocks: (1) a knowledge supervision and collaboration module (KSCM), which facilitates mutual learning between handcrafted and deep features, and (2) a feature fusion and contribution assignment module (FFCS), which enhances channel attention. The proposed model has been tested on three OpenSARShip categories and seven FUSARShip categories using both deep and shallow versions of AlexNet, VGGNet, ResNet and DenseNet. The results show that handcrafted features improve the performance of all models, with Gabor features showing the most significant average gain of features of 3.4\%. Shallow networks (VGG-11 and ResNet-19) exhibit the largest improvement over their baseline counterparts following feature fusion.

 A ship classification framework that incorporates non-stationary information of SAR targets into DL models is introduced in \cite{10671463}. By this approach, the data undergo the so-called ``second-order synchrosqueezing transform'' (SST), which allows the ship characteristics to be represented in the time–frequency domain. Then, the HOG features are extracted, thus reducing the dimensionality and providing discriminative inputs. These features are then fused with CNN representations, with ResNet identified as the most effective backbone. Experiments on 3 classes of OpenSARShip and FUSAR-Ship show clear performance gains, with accuracies of 82\% and 90\%, respectively. This study highlights the value of non-stationary signal characteristics for improving robustness and generalization in SAR ship classification.

In \cite{10578024}, a multidimensional feature joint learning framework (FJL-Framework) is introduced, designed to overcome the limitations of simple feature concatenation. The framework consists of three modules: (1) a subspace-level feature selection model with geometric constraints (SGC-GA) to refine pattern features, (2) a global–local interaction model (GFIC) to enhance deep feature learning, and (3) a joint learning model (MFJL) to fuse both types of features. This approach consistently achieves accuracies of 79.0\% and 87.6\% for 3-class OpenSARShip and 7-class FuSARShip respectively. By selectively integrating geometric, pattern, and deep features, the FJL-Framework demonstrates the importance of conflict-aware joint learning for robust SAR target recognition.

In \cite{10641086}, the use of Bayesian CNNs (BCNNs) for SAR ship classification is investigated, comparing them to deterministic CNNs. Using three classes from OpenSARShip, the study shows that a Bayesian variant of LeNet-5 (BLeNet-5) improves accuracy by 1.6\% over LeNet-5, nearly matching the performance of VGG-16 while requiring far less computation. Beyond accuracy, BCNNs provide calibrated uncertainty estimates, decomposing them into aleatoric and epistemic components. The results suggest that Bayesian models can deliver more reliable predictions and offer useful confidence measures, highlighting their potential for robust SAR ship classification in limited-data settings.

  The effect of combining handcrafted features with DL features is investigated in \cite{rs13112091} by considering four handcrafted features (HOG, naive geometric features (NGF), local radar cross section (LRCS), and principal axis features (PAFs)) in conjunction with four DL models (AlexNet, VGGNet, ResNet, and DenseNet). The authors analyze different feature injection locations and techniques, recommending injection in the final fully-connected layer with dimension unification, weighted concatenation and feature normalization (DUW-Cat-FN) for optimal results. They use three classes (bulk carrier, container ship, and tanker) from OpenSARShip and seven classes (bulk carrier, container ship, fishing, tanker, general cargo, other cargo, and other) from FUSARShip. The results reveal that injecting handcrafted features into a DL model has the potential to improve performance by up to 6\%, thus highlighting the importance of handcrafted features. In \cite{10446741}, a novel contrastive learning-based method is proposed, leveraging the fusion of handcrafted features with SAR images. It employs data augmentation techniques on both SAR images, and handcrafted features, applying distinct augmentations to each data type. Subsequently, the model calculates three types of losses during deep training to facilitate learning of two primary pretext tasks and a cluster-level task. The proposed model, named DCPNet, achieves accuracies of 73.66\% and 87.94\% for 3-class and 7-class classifications on the OpenSARship and FUSAR-Ship datasets, respectively. These results suggest the potential effectiveness of contrastive learning for ship classification tasks.

Images in OpenSARship contain information such as the length and breadth of the vessel. The ship classification performance of a DL model is enhanced in \cite{electronics10101169} by combining the outputs of a CNN and a K-nearest neighbor model (KNN), where the latter classifies new data based on the majority class of its nearest neighbors. KNN is trained to classify ships based on length and breadth taken from AIS data, while the CNN uses the image data. Three classes (cargo, tanker, and others) from OpenSARship dataset are used. The model combined with KNN showed superior performance compared to the CNN model alone.

Since a ship comprises various sections, the authors of \cite{10027936} first employ a minimum bounding rectangle (MBR) to locate the ship in the image, then subdivide the bounding box into the bow, midship, and stern sections and extract the features using two models based on the VGG network. One model leverages the data from the three sections to capture local features, while the other analyzes the entire MBR to capture global features. Finally, feature fusion is employed to classify the ships into two categories: military and civilian. The method, applied to high-resolution images captured by GaoFen-3, achieves an accuracy of 96\%.

\subsection{Use of SAR properties}

Maritime monitoring through SAR imagery represents a unique challenge in case of moving ships. Indeed, motion can cause the ship to appear shifted within the SAR image with respect to its actual geographical position. This information can be exploited proficiently for classification purposes. A complex-valued CNN (CV-MotionNet) that considers both amplitude and phase information from single-look complex SAR images is proposed in \cite{9554116}. The experiments were conducted on two datasets: a five-class simulated dataset and a three-class real-world \mbox{GaoFen-3} dataset containing cargo ships, tankers, and other categories. The results demonstrate the model's effectiveness in classifying SAR targets.

\begin{figure}[ht]
\centering
\footnotesize
\resizebox{0.9\linewidth}{!}{%
\begin{tikzpicture}[
  node distance = 6mm and 6mm,
  every node/.style={draw, rounded corners, thick, fill=white, font=\footnotesize, align=left, text width=66mm, inner sep=3pt},
  title/.style={draw, thick, rounded corners, fill=gray!20, font=\bfseries\footnotesize, text width=72mm, align=center},
  arrow/.style={-{Stealth[length=2mm]}, thick}
]

\node (root) [title] {Use of SAR Properties in Ship Classification};

\node (motion) [below=of root] {%
\textbf{Motion \& Phase–Physics (Complex SAR)}\\
\scriptsize Complex-valued CNN for motion cues \cite{9554116};\\
Phase-aware network (FDC--TA--DSN) with complex SAR \cite{10890957}
};

\node (pol) [below=of motion] {%
\textbf{Polarimetric Fusion (VV/VH)}\\
\scriptsize SE--LPN--DPFF dual-pol fusion \cite{9568903}; VGG+bilinear dual-pol \cite{9885840};\\
GBCNN w/ grouped bilinear pooling \cite{9782459,10149595};\\
DPIG-Net (PCCAF, DRDLF) \cite{rs15082138}; smoothed pol. fusion \cite{10440349};\\
Dual-branch ConvNeXt w/ pseudo-RGB pol./texture \cite{10457854};\\
Polarimetric interaction (CPINet, MO-SE, FBC) \cite{rs16183479};\\
Pol.+geometric handcrafted fusion (PFGFE) \cite{ZHANG2022108365}
};

\node (signal) [below=of pol] {%
\textbf{Scattering/Signal Conditioning}\\
\scriptsize Multiscale global scattering association (MGSFA-Net) \cite{10412205};\\
Sidelobe suppression (preprocessing) \cite{9455304}
};

\node (res) [below=of signal] {%
\textbf{Resolution \& Regime Handling}\\
\scriptsize Super-resolution $\rightarrow$ classify (TSX pretrain, OpenSARShip) \cite{8897831};\\
Siamese + detection; dual-pol inputs \cite{9172933};\\
Large-class dual-pol fusion (8-class) \cite{9387401}
};

\draw[arrow] (root.south) -- (motion.north);
\draw[arrow] (motion.south) -- (pol.north);
\draw[arrow] (pol.south) -- (signal.north);
\draw[arrow] (signal.south) -- (res.north);

\end{tikzpicture}%
}
\caption{Overview of methods that explicitly exploit SAR imaging properties for ship classification. Approaches are grouped into four categories: (i) motion and phase information from complex-valued data, (ii) polarimetric fusion of VV/VH channels, (iii) scattering- and signal-conditioning techniques, and (iv) resolution- and regime-specific strategies. Representative studies for each category are indicated by citation keys.}

\label{fig:sar_properties_condensed}
\end{figure}

Sidelobes, faint echoes in SAR images caused by the finite length of the radar pulse, can hinder the performance of DL models for ship classification. The effect of sidelobe suppression is investigated in \cite{9455304} using a simple CNN model on ten classes from FUSARShip (anti-pollution, bulk carrier, container, dredger, fishing, cargo, law enforce, passenger, oil tanker, and port tender). The results show that suppressing the sidelobes improves the model performance, achieving up to 98\% accuracy.

The authors of \cite{9568903} investigate whether using information from two different polarizations, VV and VH, can improve the model performance by capturing more useful features. They propose a novel network called squeeze-and-excitation Laplacian pyramid network with dual-polarization feature fusion (SE-LPN-DPFF). This first merges features extracted from both polarizations, then uses squeeze and excitation blocks to enhance feature representation by selectively emphasizing informative features. Finally, the features are fed into a Laplacian pyramid network based on residual blocks, which extracts information from different resolutions and performs classification ensuring that the model is trained on this fine-grained information. Two datasets from OpenSARship are used, one with three categories (bulk carriers, container ships, and tankers) and another with six categories (bulk carriers, cargo, container ships, fishing, general cargo and tankers). To ensure a balanced dataset, the authors use 70\% of the data from the category with the fewest samples as the training set and the remaining data for testing. They compare the performance of SE-LPN-DPFF to 35 other DL classifiers. The results show that SE-LPN-DPFF outperforms all other models. 

 A different approach is proposed in \cite{9885840}, using a dual-polarized network based on a pretrained VGG architecture. This network learns representations from both VV and VH polarizations and then applies bilinear pooling. The authors train the model on a three-class dataset from OpenSARShip (tanker, container ship, and bulk carrier), where VV and VH images are paired for training. The results show that the model outperforms state-of-the-art models, achieving an accuracy of 88\%. A novel methodology, named the group bilinear convolutional neural network (GBCNN), is proposed in \cite{9782459, 10149595}, combining a loss function and bilinear pooling, based on dense blocks. Grouped VV and VH images are used to train two parallel models. The extracted features are then passed through bilinear pooling. The resulting features from both pooling layers are sent to the loss function, along with a third feature created by merging the pooling outputs. Using three- and five-class datasets from OpenSARShip (tanker, container ship, bulk carrier, cargo, general cargo), with data augmentation techniques such as flipping, rotation, translation, and Gaussian noise, the GBCNN achieved accuracy scores of 88\% and 67\% for three and five classes, respectively.

Recent work has begun to explore the use of complex-valued SAR data for ship classification, leveraging both amplitude and phase information rather than relying on intensity alone. The authors of \cite{10890957} introduce FDC-TA-DSN, a deep SAR network that combines four-dimensional dynamic convolution (FDC) to suppress speckle noise with a triple-attention (TA) mechanism to filter sea clutter and emphasize discriminative features. Phase information is incorporated through joint time–frequency analysis and autoencoder-based feature learning, and fused with amplitude features to enhance classification. To support this effort, a new high-resolution dataset, ComplexSAR\_Ship, was constructed from Gaofen-3 imagery, covering 17 ship categories. Experiments show that FDC-TA-DSN consistently outperforms standard CNNs and DSN baselines, confirming the value of phase information and complex-valued features for more robust ship classification.

In \cite{ZHANG2022108365}, polarization is investigated in conjunction with handcrafted features. A novel polarization fusion-based geometric features network (PFGFE) is proposed, utilizing geometric features alongside VH-polarized, VV-polarized and a combined VV-VH image. PFGFE employs three VGG network-based CNNs for feature extraction. A global feature self-attention mechanism is then applied to the output of each CNN. Subsequently, the extracted features are merged, and their classification probabilities are concatenated with corresponding coefficients. Geometric features are processed using LSTM to capture the importance of long-term dependencies. Finally, the outputs from the CNNs and LSTM are concatenated, and this final representation is used for classification. The authors evaluate PFGFE on three-class and six-class datasets from OpenSARShip (respectively, bulk carrier, container ship, and tanker, and bulk carrier, cargo, container ship, fishing, general cargo, and tanker). The results show accuracies of 79.84\% and 56.83\% respectively.

Building on the idea that exploiting dual polarization information can enhance performance, the authors of \cite{rs15082138} propose a novel dual-polarization information-guided network (DPIG-Net) that leverages a polarization channel cross-attention framework (PCCAF) and a dilated residual dense learning framework (DRDLF) for effective feature extraction and fusion. It evaluates DPIG-Net on the OpenSARShip dataset using VH and VV polarizations for three (bulk carrier, container ship, tanker) and six classes (adding cargo, fishing, and general cargo), achieving accuracies of 81.28\% and 58.68\%, respectively. Smoothed polarization fusion, coupled with attention modules such as multiscale feature fusion and dynamic gating feature fusion, has been shown to improve model accuracy for ship classification. This approach is adopted in \cite{10440349} on  OpenSARShip, achieving accuracies of 87.13\% and 65.97\% for three and six classes, respectively. These results suggest that smoothed polarization fusion can be more effective than coherence fusion \cite{rs15082138}. In \cite{10457854}, a novel dual branch deep network for ship classification is proposed and tested on OpenSARship. The network leverages VH and VV polarizations and comprises three modules: (1) an image construction module that generates two pseudo-RGB images representing polarimetric and texture features, (2) a feature extraction module that employs two pretrained ConvNeXt networks to extract features from the constructed images independently and (3) a feature fusion and classification module that combines the extracted features and performs classification. This network achieves accuracies of 87.62\% and 64.26\% for 3-class and 6-class classification tasks, respectively.

 A Multiscale Global Scattering Feature Association Network (MGSFA-Net) is proposed in \cite{10412205}. It integrates scattering features with deep features for SAR ship classification. After fine segmentation to isolate ship targets, scattering centers (SCs) are extracted and modeled as local graphs, which are then associated through the SCFA module and enhanced with a multiscale fusion module. In parallel, deep features are extracted using a multikernel module, and both representations are fused by weighted integration. Experiments on FUSAR-Ship and OpenSARShip demonstrate that MGSFA-Net achieves accuracies of 85.3\% (three-class FuSARShip) and 85.7\% (three-class OpenSARShip), outperforming other feature-fusion methods. The results show that incorporating scattering characteristics improves robustness and interpretability, particularly in few-shot scenarios.

 A Cross-Polarimetric Interaction Network (CPINet) is proposed in \cite{rs16183479} to exploit the complementary information in dual-polarized SAR images. The network employs a multiscale deep feature extraction framework and introduces a mixed-order squeeze–excitation (MO-SE) attention module, where first- and second-order statistics from one polarization guide the learning of the other. To further enhance feature fusion, factorized bilinear coding (FBC) is adopted, which reduces the parameter size while suppressing noise. An adaptive loss balancing strategy based on GradNorm is also applied to optimize multitask training. Evaluated on three- and five-class subsets of OpenSARShip, CPINet consistently outperforms both single-pol baselines and prior dual-pol methods, demonstrating the effectiveness of polarimetric feature interaction for robust ship classification.

The limited spatial resolution of SAR imagery presents a significant challenge for ship classification tasks, particularly for small vessels. These ships often occupy only a few pixels within the image, making it difficult to extract the features needed for an accurate classification. To address this, the authors of \cite{8897831} propose a joint convolutional network to enhance the spatial resolution before classification. It uses TerraSAR-X images for pretraining and three classes from OpenSARship for training (bulk carrier, container, tanker). The data are first fed into a generative model for upscaling, and then classified using a dense blocks-based convolutional network. It also compares the performance of models trained on VV and VH polarizations. The results show an increase in performance compared to simpler networks, achieving classification accuracies of 81.02\% and 79.12\% for VH and VV, respectively.

Methods such as the siamese networks can be helpful in overcoming the challenge of scarce and imbalanced data. Siamese networks focus on learning a similarity function by training two networks that share weights, allowing for better learning of data representations. In \cite{9172933}, a novel loss function with a siamese network architecture are proposed. Two inception blocks are used on different polarizations (VV and VH) for four classes from OpenSARship (cargo, tanker, tug, and other). The images are first passed through a target recognition algorithm (RCNN) to crop the targets. Then, the cropped images are fed into the inception blocks, the features are extracted and the loss is calculated. The results indicate that siamese networks with target detection outperform models without target detection, achieving an accuracy of 87.04\%. 

In ship classification tasks, researchers typically use datasets with three or four categories. This is because DL models require a large amount of data for training, and maritime data are inherently imbalanced, making it challenging to achieve good performance with more classes. However, a study in \cite{9387401} demonstrates the potential of feature fusion for handling a larger number of categories. It employs an eight-category dataset from OpenSARShip (cargo, tanker, carrier, container, fishing, dredger, tug, and passenger). A VGG-based feature extractor is used to simultaneously extract feature maps from both VV and VH polarizations. These feature maps are then merged. Finally, a novel hybrid loss function is applied to reduce intra-class distances and increase inter-class distances, leading to improved classification performance. The results, with a mean average precision of 82\%, support the effectiveness of this approach.

\subsection{Transfer Learning and Domain Adaptation}

\begin{figure}[ht]
\centering
\footnotesize
\resizebox{0.9\linewidth}{!}{%
\begin{tikzpicture}[
  node distance = 6mm and 6mm,
  title/.style={draw, rounded corners, thick, fill=gray!20,
                font=\bfseries\footnotesize, align=center,
                text width=70mm, inner sep=4pt},
  box/.style={draw, rounded corners, thick, fill=white,
              font=\footnotesize, align=left,
              text width=70mm, inner sep=3pt},
  arrow/.style={-{Stealth[length=2mm]}, thick}
]

\node (root) [title] {Transfer Learning and Domain Adaptation in SAR Ship Classification};

\node (finetune) [box, below=of root] {%
\textbf{Fine-Tuning Pre-trained Models}\\
\scriptsize VGG-16 fine-tuning (f1 = 97.8\%) \cite{s18092929};\\
DenseNet-121 + augmentation (98\%) \cite{s19010063};\\
Impact of incident angle variation \cite{8899277}
};

\node (domain) [box, below=of finetune] {%
\textbf{Domain Adaptation}\\
\scriptsize DSAN++ (optical $\rightarrow$ SAR) \cite{9893336};\\
Cross-dataset adaptation (OpenSARShip $\leftrightarrow$ FUSARShip) \cite{10547709};\\
Domain-specific vs.~non-domain-specific pretraining \cite{rs14235986}
};

\node (aug) [box, below=of domain] {%
\textbf{Data Augmentation \& Generative Models}\\
\scriptsize Augmentation + handcrafted features \cite{RELEKAR20214594};\\
Conditional GANs for high-res SAR \cite{9180260};\\
Generative domain adaptation \cite{10497597};\\
Attention-Dense CycleGAN (OPT2SAR) \cite{10246308}
};

\draw[arrow] (root.south) -- (finetune.north);
\draw[arrow] (finetune.south) -- (domain.north);
\draw[arrow] (domain.south) -- (aug.north);

\end{tikzpicture}%
}
\caption{Overview of transfer learning strategies for SAR ship classification. The approaches are grouped into three categories: (i) fine-tuning of pre-trained models, (ii) domain adaptation across modalities or datasets, and (iii) data augmentation and generative models to mitigate data scarcity. Representative studies are indicated by citation keys.}
\label{fig:finetuning_transfer}
\end{figure}

High resolution images (e.g. from CosmoSkyMed mission, level 1A HIMAGE single-look complex slant (SCS) in the X-band, single HH or VV polarization) can be exploited to test the effectiveness of pre-trained models on datasets that have a limited number of classes (e.g. three). In \cite{s18092929} the authors follow this approach and conclude that fine-tuning a VGG-16 network can lead to \mbox{F1-scores} of up to 97.8\%. This shows the usefulness of DL techniques for small-sized datasets (like usual SAR imagery datasets).

In \cite{s19010063}, the authors demonstrate that leveraging transfer learning with pre-trained models (based on DenseNet-121 architecture), combined with effective data augmentation strategies, can significantly improve the classification accuracy, even when working with small-sized training datasets. Stripmap input images were sourced from TerraSAR-X mission and have been manually annotated by experts using AIS data. Only three classes are employed (bulk carrier, container ship, and oil tanker), and data augmentation techniques are applied (flipping, brightening, sharpening, random cropping, and rotating). Despite the tiny size of the training dataset, the authors achieved a 98\% accuracy.

In \cite{8899277}, the authors examine the effect of varying incident angles on model performance. They use a dataset containing three classes of images taken from Cosmo-SkyMed and \mbox{GaoFen-3}, encompassing incident angles ranging from 20 to 60 degrees. The data were divided into three overlapping subsets: one containing the entire dataset, a second one containing data with low incident angles (less than 48 degrees), and a third one containing data with high incident angles (more than 45 degrees).
%
All three datasets were tested on four pre-trained DL models: VGG-16, VGG-19, ResNet-50, and ResNet-101. The results indicate that incorporating data from all available angles enhances the models' predictive accuracy.

Data scarcity can be addressed by training the model on domain-specific data, such as optical images of ships, then applying domain adaptation to SAR data. In \cite{9893336}, DSAN++, a novel method for domain adaptation, is proposed. This first involves training on optical images and then transferring the learned knowledge to SAR data. DSAN++ utilizes two ResNet networks for feature extraction. The extracted features undergo feature fusion, followed by the application of a subdomain adaptation network to learn transferable representations. To capture both local and global distributions, shallow and deep ResNet models (ResNet-18 and ResNet-50) are integrated with the convolutional block attention module. The experiments were conducted by training the model on an optical remote sensing (ORS) dataset obtained from Google Earth, with four classes (cargo, bulk carrier, container ship, and oil tanker). Subsequently, domain adaptation was performed on two datasets: (1) GF-SAR with three classes (bulk carrier, container ship, and oil tanker) collected from FUSAR and GaoFen, and (2) a three-class dataset (cargo, container ship, and oil tanker) derived from TerraSAR-X. The results demonstrate that DSAN++ leads to improved performance in subdomain adaptation. In \cite{10547709}, the authors apply domain adaptation from OpenSARship to FuSARShip and vice versa on multiple scales resulting in better performance for both datasets as compared to the one obtained by training and testing on a single dataset. In \cite{rs14235986}, the authors compare the fine-tuning performance of models pre-trained on domain-specific versus non-domain-specific datasets using CNNs. The methodology was evaluated on three-class targets from VV polarized TerraSAR-X (carrier, container, and oil tanker), and four-class targets  from FUSARShip (cargo, bulk carrier, container, and oil tanker). The results suggest that shallow models with feature refinement can achieve better performance when fine-tuned on domain-specific models. The authors investigate the pre-training effects of ImageNet and an ORS dataset with data augmentation techniques (scaling, flipping, and rotating). Whereas ImageNet, due to its large size, yields better performances on deep networks, the ORS dataset enhanced the performance of shallow networks, indicating that shallow models are more effective with less data.

\subsubsection{Data Augmentation \& GANs}

Techniques like data augmentation, when combined with transfer learning, can address data-related problems. In \cite{RELEKAR20214594}, the authors demonstrate that using transfer learning with data augmentation and manual features can improve the performance of a DL model. They use ground range detected SAR data acquired in interferometric wide-swath mode from OpenSARShip. The data consist of two channels, with VH and VV polarizations, stacked together. The study focuses on classifying ships into four categories: bulk carrier, container ship, fishing vessel, and tanker. The authors use five data augmentation techniques (horizontal flipping, vertical flipping, rotation by 90°, rotation by 270°, and Gaussian noise injection) and incorporate 14 handcrafted features. The effectiveness of these handcrafted features combined with DL features are compared across three machine learning and four DL models.The ML models are Radial Basis Function (RBF)-SVM with two stage feature selection, SVM classifier combination, SVM one-vs-rest; the DL models are CNN, AlexNet, VGG-19, and ResNet50. Additionally, the authors present a methodology for matching AIS and SAR data and test the performance on real data collected from three different locations. Their results show that the pretrained ResNet50 model, when fine-tuned on SAR images and handcrafted features, achieves an accuracy of 82.5\%.

One of the possible DL-based solutions to the data scarcity problem in optical images is represented by generative models and can also be applied to SAR images. These models learn data representations and produce images that resemble the originals, effectively increasing the available data. In \cite{9180260}, the authors propose a novel conditional generative adversarial network method (GAN) that generates high-quality SAR images, improving ship classification performance. The method consists of two parts: (1) a residual convolutional block is integrated with the generator to enhance the detail of generated images, and (2) the generated images are merged with real images and sent to a discriminator for two tasks: distinguishing real from fake images and classifying them into three classes. The authors use a three-class dataset (cargo, container, tanker) from 
GaoFen-3. The results indicate that the GAN is capable of generating high-resolution images that, coupled with the real images, enhance the classification performance. 

A generative model-based domain adaptation framework for ship classification on a \mbox{5-class} FUSARShip dataset is proposed in \cite{10497597}. This framework leverages domain adaptation by transferring knowledge from the optical images of the FGSCR42 dataset. In \cite{10246308}, the Attention-Dense CycleGAN (ADCG) is introduced to improve recognition  by leveraging transfer learning from optical to SAR domains (OPT2SAR). ADCG enhances CycleGAN with a dense connection module (DCM) and a convolutional block attention module (CBAM). Using FUSARShip and the FGSCR optical dataset, ADCG achieves superior performance, evaluated via Fréchet inception distance (FID) and kernel inception distance (KID). It demonstrates a 6\% improvement in recognition accuracy for SAR ship classification networks, thus validating its effectiveness in addressing class imbalance and sample scarcity through effective transfer learning.

\section{Findings}\label{sec5}

\subsection{Datasets}
The literature explores two public datasets, FUSARShip and OpenSARShip, and some additional, distinct data sources for SAR ship classification. More detailed information is provided below.
\subsubsection{OpenSARShip}
The OpenSARship dataset is a publicly available resource designed to support research in ship classification using SAR imagery. This dataset provides a collection of pre-processed SAR image chips specifically focused on ship detection and classification tasks. The current version contains 34528 vessel instances. The data are collected using two products from ESA's Sentinel-1 repository (single look complex, and ground range detected) in interferometric wide swath mode. The dataset consists of 16 ship typologies, among which classes like Cargo, Tanker, Fishing, Tug, Dredging can be found. However, the dataset suffers from class imbalance, with 61\% data belonging to Cargo category. Three classes represent 95\% of data, while the remaining 13 classes only represent a 5\% of the data provided. This imbalance can lead any model to bias toward the majority class. The literature shows that researchers often address this problem by only using the 3 to 5 most populated classes, e.g., cargo, tanker, tug, fishing, and dredging \cite{s18093039, Zhu_2020, 10147843, 9146990, 9564861, 9445223, 10189816}, or by utilizing subclasses within the cargo category, e.g., general cargo, container ship, bulk carrier \cite{9199258, 9553116, 9805965}. OpenSARship data incorporate polarization information, VH and VV, which can be leveraged to create fusion networks \cite{app12146866, rs15112904, 9568903, 9885840, 9782459, 10149595, ZHANG2022108365, rs15082138, 10457854}. The dataset's main weaknesses such as low spatial resolution (20 m/pixel) and class imbalance, limit the performance of any DL model. Nevertheless, if these limitations are addressed, OpenSARship serves as a valuable starting point for researchers to explore their methodologies before acquiring commercial SAR imagery.
\subsubsection{FUSARShip}
FUSARShip provides a valuable resource for researchers in SAR ship classification. This dataset uses high-resolution SAR images captured by China's GaoFen-3 (GF-3) satellite, specifically designed for ocean remote sensing and marine monitoring tasks. \mbox{FUSARShip} is constructed from a collection of image chips extracted from over 126 GF-3 scenes, encompassing 15 primary ship categories
and 98 subcategories. Each ship instance has a size of $512\times 512$ pixels, with a pixel size of 0.5 meters. Similar to OpenSARship, FUSARShip presents a challenge due to class imbalance. Three sub-categories constitute approximately an 81\% of the data, while the remaining 95 categories collectively only represent a 19\% thereof. This significant imbalance can lead to models exhibiting bias towards the overrepresented classes. Consequently, researchers often address this issue by focusing on the 3-5 most populated classes, e.g., cargo, fishing, tanker, bulk carrier \cite{10433638, rs15112917, rs15112904, 9893336, rs14235986, 10497597, 10246308} as observed in the OpenSARship dataset.
\subsubsection{Other Data Sources}
Large datasets are crucial for training a DL model. However, the imbalance data limitation in the above datasets can hinder model performance. Researchers address this challenge using two techniques. Firstly, they leverage high-quality data from various satellite missions, such as TerraSAR-X \cite{8113469, 8897831, s19010063, 9893336, rs14235986}, CosmoSkyMed \cite{s18092929}, PALSAR-2 \cite{10.1007/978-3-030-01054-6_2}, and GaoFen-3 \cite{10027936, 9554116, 8899277, 9180260}, see  Table~\ref{tab:sar_platforms}. These missions not only provide high quality data, but also allow researchers to exploit AIS to strategically select images based on specific locations and timeframes. For instance, if a researcher aims to develop a model for detecting fishing activity during a particular season in a specific region, AIS data can be analyzed to identify relevant satellite images for purchase.

Secondly, due to the cost associated with acquiring high-resolution commercial imagery, researchers often employ transfer learning techniques. This involves training a model on a large, general-purpose dataset (e.g., ImageNet), or domain specific dataset, e.g., MSTAR \cite{afMSTAROverview} (SAR images of tanks), optical remote sensing images of ships taken from Google Earth \cite{9784428} and then on a smaller, task-specific dataset (e.g., FUSARShip). Pretraining on a comprehensive dataset equips the model with the ability to extract generic features from images. Subsequent fine-tuning on the target dataset refines these features for the specific classification task.
\begin{table*}[t]
    \centering
    \caption{SAR Platforms and their resolutions}
    \label{tab:sar_platforms}
    \begin{tabular}{|c|c|c|c|}
    \hline
    \textbf{Platform} & \textbf{Band} & \textbf{Image Modes}  & \textbf{Resolution(s)} \\
    \hline
    Sentinel-1 & C & \multicolumn{1}{c|}{\begin{tabular}[c]{@{}c@{}}StripMap, \\ Interferometric Wide Swath (IW), \\ Extra Wide Swath (EW)\end{tabular}}   &  \begin{tabular}[c]{@{}l@{}}- StripMap: $5 \times 5$ m\\ - IW: $5 \times 25$ m \\ - EW: $20 \times 40$ m\end{tabular} \\
    \hline
    Cosmo-SkyMed & X & \multicolumn{1}{c|}{\begin{tabular}[c]{@{}c@{}}Spotlight, \\ StripMap, \\ Wide \end{tabular}} & \begin{tabular}[c]{@{}l@{}}- Spotlight: Up to 1 m\\ - StripMap: Up to 3 m\\ - Wide: Up to 30 m\end{tabular} \\
    \hline
    TerraSAR-X & X & \multicolumn{1}{c|}{\begin{tabular}[c]{@{}c@{}}StripMap, \\ Spotlight, \\ High Resolution Spotlight (HRS)\end{tabular}}  &  \begin{tabular}[c]{@{}l@{}}- StripMap: Up to 3 m\\ - Spotlight: Up to 0.3 m\\ - HRS: Up to 0.3 m\end{tabular}\\    \hline
    Gaofen-3 & X & \begin{tabular}[c]{@{}l@{}}StripMap\\ Spotlight\end{tabular}  &   \begin{tabular}[c]{@{}l@{}}- StripMap: Up to 2.5 m\\ - Spotlight: Up to 0.9 m\end{tabular}\\
    \hline
    PALSAR-2 & L & \begin{tabular}[c]{@{}l@{}}StripMap\\ Spotlight\end{tabular}  &   \begin{tabular}[c]{@{}l@{}}- StripMap: $3 \times 3$ m\\ - Spotlight: $1 \times 3$ m\end{tabular}\\
    \hline
    \end{tabular}
\end{table*}

\subsection{Data Augmentation}
Due to the limited availability of SAR images and the inherent data requirements of DL models, data augmentation techniques have been proposed in the literature to address this challenge. These techniques can be employed in an attempt to improve the performance of DL models for SAR ship classification.

\subsubsection{Rotation}
in the literature, rotation is widely employed to enhance the robustness of DL models for SAR ship classification \cite{RELEKAR20214594, 9782459, 10149595, rs15112917, 9146990, 8113469, rs15112904}. It involves artificially rotating the training images by various angles to simulate real-world scenarios where objects might be oriented differently relative to the SAR sensor. A common approach is to rotate images within a range of $+45$ to $-45$ degrees, as this range captures a significant portion of the potential ship orientations. This augmentation strategy helps the model learn features that are invariant to rotation, improving its generalization ability and reducing the risk of overfitting to the specific orientations present in the original training data.


\subsubsection{Flipping}
the literature commonly employs horizontal and vertical flipping data augmentation techniques \cite{app12146866, rs14235986, 10149595, s19010063, RELEKAR20214594, rs15112904, rs15112917}. These techniques generate new images by mirroring the ships along the horizontal or vertical axes. This process simulates scenarios where ships are viewed from opposite directions, effectively augmenting the dataset and potentially improving the model's ability to generalize to unseen orientations.

\subsubsection{Scaling}
DL networks typically require a fixed-size input. Image scaling, a data augmentation technique that resizes images to a standard size, addresses this requirement. Scaling can be performed by increasing or decreasing image size, with larger images demanding significantly more computational resources for processing \cite{rs14235986, rs15112904}. In SAR ship classification, a common practice is to scale images to a size of $128
\times 128$ pixels, balancing accuracy and computational efficiency.

\subsubsection{Translation}
in image processing, translation refers to the spatial transformation of an image by a specific offset in both horizontal and vertical directions. This essentially moves the entire image content without any scaling or rotation. In SAR ship classification, translation augmentation can be employed to artificially introduce variations in the training data \cite{9782459, 8113469, 9146990, 10149595}. By exposing the DL model to images with slightly shifted ship positions, translation augmentation can potentially improve the model's generalization ability. This is because the model learns to recognize ships irrespective of their small location variations within the image, making it more robust to real-world scenarios.

\subsubsection{Gaussian Filtering}
SAR images of marine scenes are often corrupted by a strong noise component. To address this challenge, Gaussian filtering is commonly employed \cite{9146990, 9782459, 10149595, RELEKAR20214594}. This technique works by averaging the intensity values of pixels within a localized neighborhood around each pixel. This process reduces the random variations in pixel intensity caused by noise, enhancing the distinction of ship features from the background noise.

\subsubsection{Cropping}
maritime SAR imagery contains a lot of information that is useless for ship classification purposes. In fact, the representation of the ship target is usually way smaller than the surrounding background, and this leads the model to learn non-informative details. As a possible countermeasure to reduce useless information, cropping, i.e. the process of extracting a region of interest containing the ship target only, is adopted. Cropping is commonly employed in literature as a data augmentation technique \cite{s19010063, 9172933, rs15112904, 8127014}.


\subsection{Handcrafted Features}
While data augmentation is often used to increase training diversity, it can also lead to overfitting, especially when applied to low-quality SAR images. In such cases, augmentation may generate additional noisy samples that confuse the model rather than improve it. Handcrafted features provide an alternative source of information that can be fused with DL features to strengthen classification robustness (see Fig.~\ref{fig:feature_fusion}).

DL models excel at automatically extracting features, yet their performance in SAR ship classification is constrained by data scarcity, imbalance, and strong speckle noise. To address these issues, researchers have integrated handcrafted features that embed prior knowledge about ship structure and radar scattering. Below, we review the most commonly used handcrafted features in the literature.

\subsubsection{Histogram of oriented gradients}
\begin{figure}[H]
    \centering
    \includegraphics[scale=0.2]{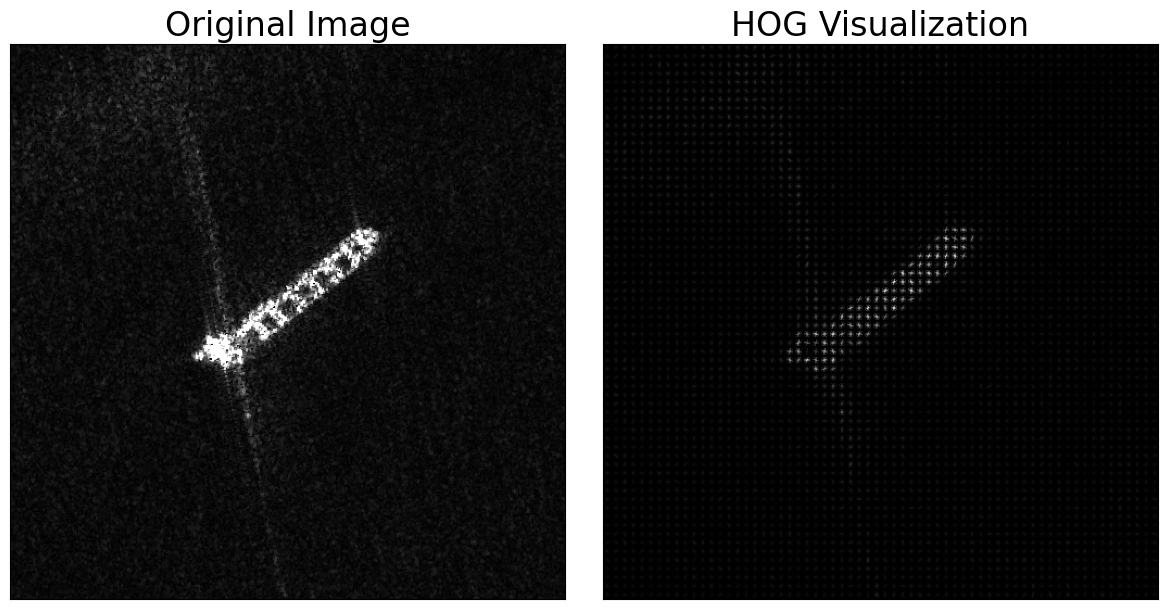}
    \caption{HOG features of a SAR image}
    \label{fig:hog_example}
\end{figure}

Histogram of Oriented Gradients (HOG, see Fig.~\ref{fig:hog_example}) captures edge orientations and local gradient structures \cite{9553192, 9445223, rs13112091}. In SAR images, where ship targets typically exhibit strong geometric edges, HOG can highlight hull and superstructure outlines. However, speckle noise can degrade gradient estimation, making HOG alone unreliable. Recent works therefore combine HOG with DL features, where it contributes complementary edge information that can guide the network’s learning of ship contours.

\subsubsection{Local Radar Cross-Section}
Local Radar Cross-Section (RCS) features describe scattering intensities within specific image regions. These features directly capture radar-domain priors that are not always learned effectively by DL models. When fused with learned features, local RCS provides physical interpretability and helps reduce confusion between ships with similar shapes but different scattering patterns \cite{rs13112091}. Nonetheless, RCS features are sensitive to resolution and viewing geometry, which may limit generalization across datasets.

\subsubsection{Geometric Features}
\begin{figure}[H]
    \centering
    \includegraphics[scale=0.2]{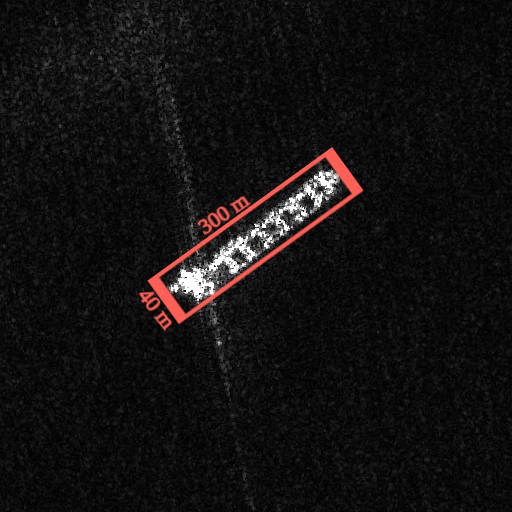}
    \caption{Bounding Box of a ship target in a SAR image chip.}
    \label{fig:geometric}
\end{figure}

Geometric features (such as length, width or related combinations) can be proficiently exploited for classification. During training with annotated datasets, these features can be drawn from the AIS information \cite{electronics10101169}. During operational classification, image processing algorithms can be applied to segment ship targets (e.g. through bounding box fit) and, provided that the resolution is known, estimate the geometric variables of interest \cite{10027936}, as shown in Fig.~\ref{fig:geometric}. Length and width can be used to extract additional features \cite{rs13112091}, such as area, perimeter, aspect ratio, shape etc. Adding this information can help DL models to improve the ship classes prediction.


\noindent Overall, handcrafted features act as complementary signals rather than stand-alone solutions. HOG captures edge structure but is noise-sensitive, RCS injects radar-specific priors but depends on resolution, and geometric descriptors are highly discriminative yet prone to segmentation errors. Their effectiveness lies in feature fusion with DL models, especially under data scarcity and imbalance, where handcrafted cues can provide stability and interpretability.


\subsection{Training Techniques}
\label{sec:training}
Although handcrafted features can be helpful, they require high-resolution images for effective extraction, which is often not the case, as seen in the OpenSARShip dataset. To address this challenge, researchers have explored novel DL techniques, further explained below.

A DL model takes input in the form of an image, learns the features, and then predicts the class of the image. However, different training strategies can be attempted to achieve promising classification performances. The techniques used in the literature for better classification are described in this section.

\subsubsection{Feature Fusion}
\begin{figure}[H]
    \centering
    \includegraphics[scale=0.2]{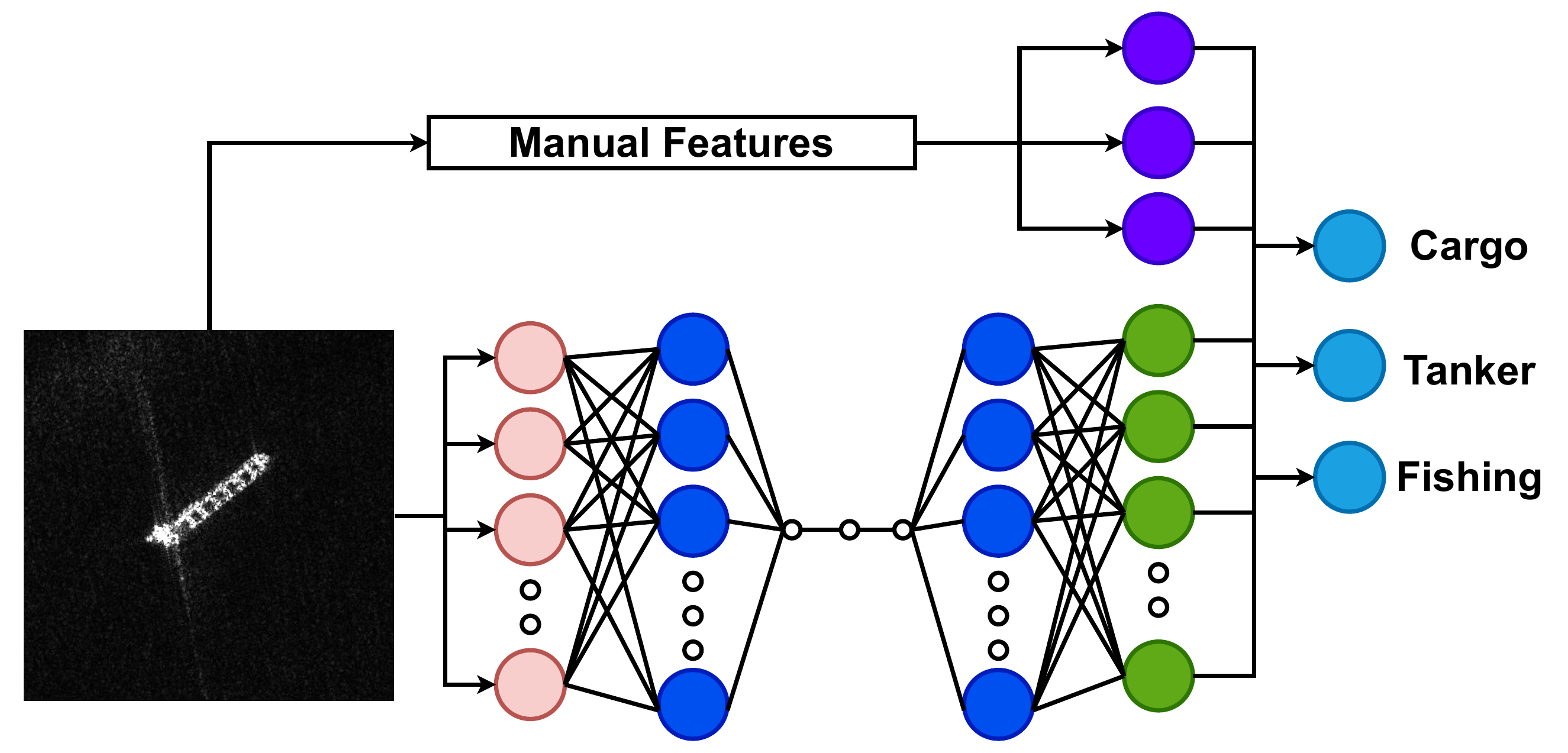}
    \caption{Feature Fusion}
    \label{fig:feature_fusion}
\end{figure}

a DL model trains on the features that it extracts, but when it comes to SAR ship classification, the challenge of data scarcity becomes a hurdle on the way to better results. To help the model understand the data well, several researchers propose to add handcrafted features to the DL model \cite{9553192, 9445223, 10487903, 10189816, 10446741}. Deep and handcrafted features are merged and the model classifies on this merged information, thus obtaining a better classification performance \cite{rs13112091}. A merging strategy is schematized in Fig.~\ref{fig:feature_fusion}. 

Feature fusion is a useful technique; however, it requires manual effort, as there is no existing pipeline to extract these features automatically. Additionally, there are no standardized features that consistently enhance classification performance. It is important to note that the choice of features is resource- and time-intensive and does not always lead to better results.

\subsubsection{Polarization fusion}
\begin{figure}[b]
    \centering
    \includegraphics[scale=0.18]{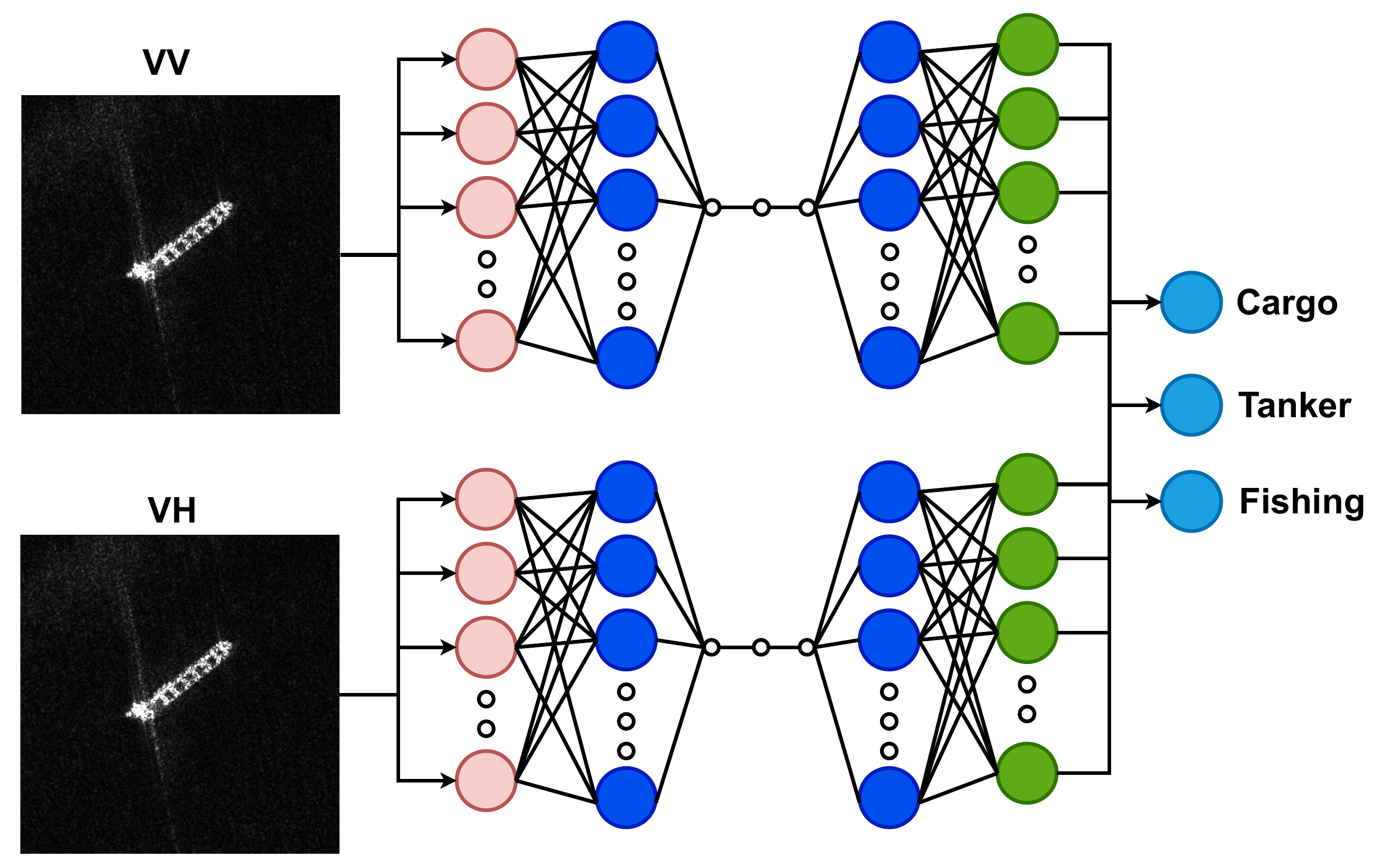}
    \caption{Polarization Fusion (simplified version from \cite{9885840})}
    \label{fig:pol_fusion}
\end{figure}

SAR images can be acquired in four transmit-receive polarizations: VV, VH, HH, and HV. While most studies in the literature have focused solely on VV or VH polarizations, it is important to note that each polarization provides distinct information about the target. Utilizing multiple polarizations 
has been shown to improve classification performance. This is because combining information from different polarizations allows the model to learn more discriminative features, leading to enhanced classification accuracy. A common approach in the literature involves creating image pairs 
and training two separate DL networks simultaneously, one for each polarization \cite{ZHANG2022108365, 9568903, 9885840, rs15082138, 9172933}. As illustrated in Fig.~\ref{fig:pol_fusion}, the features extracted from these networks are then merged for final classification.

Exploiting the polarization information is indeed a useful technique that provides the model with more data to learn from; however, this is not always readily available. From a cost perspective, researchers must purchase two images instead of one, significantly increasing the acquisition cost. For public datasets, polarization is available only in the OpenSARShip dataset and not in the FUSARShip dataset, limiting experimentation. Furthermore, training two models doubles the training cost and increases resource requirements.

\subsubsection{Metric Learning}
\begin{figure}[H]
    \centering
    \includegraphics[scale=0.2]{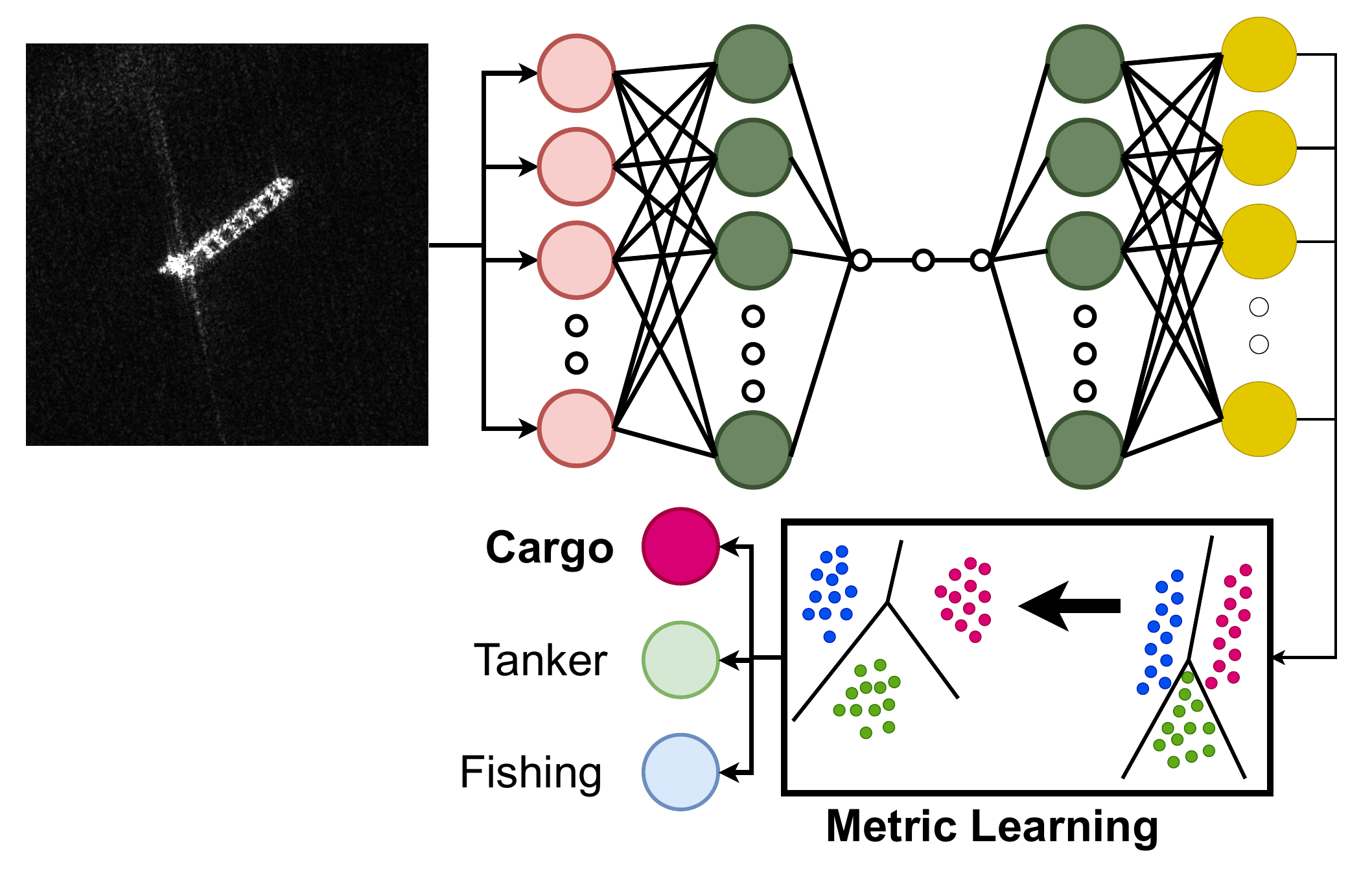}
    \caption{Metric Learning}
    \label{fig:metric_learning}
\end{figure}

DL models rely on feature extraction, which involves iteratively adjusting the model's weights to optimize the extracted features. Ideally, for effective classification, the features corresponding to each class should cluster tightly in a low-dimensional space, while remaining well-separated from features of other classes. However, standard training procedures may not inherently prioritize this inter-class separation. In such cases, metric learning techniques are employed to explicitly address this challenge \cite{9199258, 10433638, 9146990}. Metric learning aims to embed the extracted features into a low-dimensional space while maximizing the similarity between features belonging to the same class and minimizing the similarity between features from different classes. This is often achieved by utilizing distance metrics like cosine similarity or Euclidean distance, as illustrated in Fig.~\ref{fig:metric_learning}. Metric learning helps the model clearly distinguish between features of different classes, leading to superior performance. However, since the data is often naturally imbalanced, metric learning may be prone to overfitting. Therefore, it is crucial to balance the data beforehand or account for the imbalance when applying metric learning.

\subsubsection{Multi-Scale/Pyramid Training}
\begin{figure}[b]
    \centering
    \includegraphics[scale=0.2]{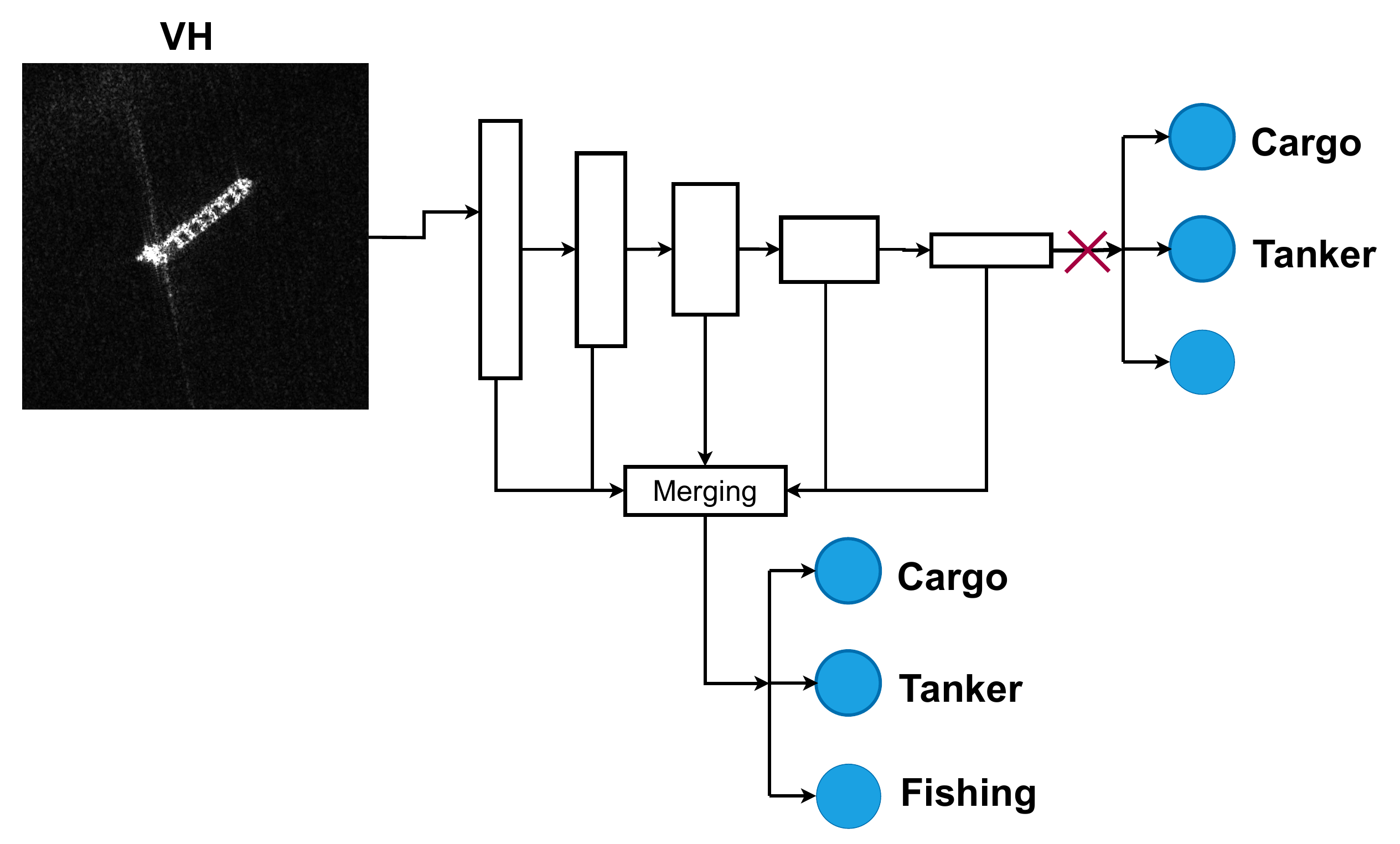}
    \caption{Pyramid networks}
    \label{fig:pyramid_net}
\end{figure}

A DL model consists of several layers; each layer learns different scaled features, and the model makes the prediction based on the features of the last convolution layers \cite{10242360, 10239480, 9564861, 9553116, 10315181, 9568903, 10937719}. In pyramid training, see Fig.~\ref{fig:pyramid_net}, the features from all the layers are directly used in decision-making by merging them using concatenation or other statistical measures (e.g. average, addition etc.). 
Although the multiscale technique is effective at introducing features from different layers, providing the model with more data to learn from, deciding which layers to use is a critical aspect. Including all layers may introduce irrelevant information, leading the model to learn unhelpful patterns. Additionally, the method of merging features significantly impacts performance. For instance, concatenating features increases the feature size, while summing features may result in the loss of important information. Features from different scales often have varying sizes, which must be upsampled or downsampled for merging. Therefore, adopting an appropriate approach to align feature sizes is crucial for achieving better classification performance.

\subsubsection{Transfer-Learning and Domain Adaptation}
\begin{figure}[h]
    \centering
    \includegraphics[scale=0.20]{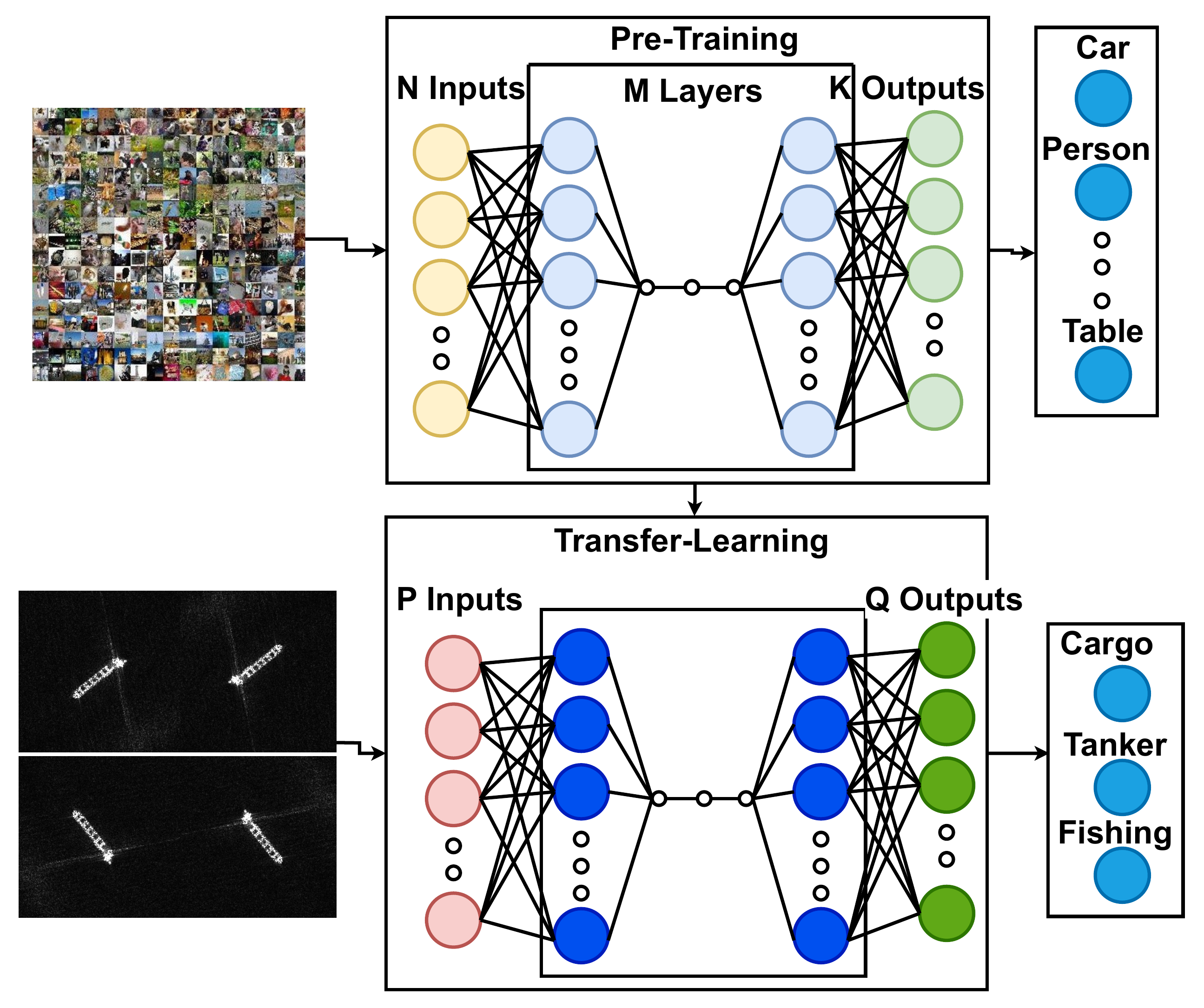}
    \caption{Transfer Learning}
    \label{fig:transfer_learn}
\end{figure}

SAR data are expensive and difficult to access, while DL models require large amounts of data. A deep neural network learns how to extract features and then applies classification to these learned features. Transfer learning can be beneficial in such scenarios. It allows leveraging other datasets to help a network learn feature extraction and then fine-tune the network with the SAR data. Transfer learning can be applied in various ways, see Fig.~\ref{fig:transfer_learn}:
\begin{itemize}
    \item[-] Using an optical image dataset, such as ImageNet, to pre-train a network, then fine-tuning on the SAR dataset \cite{s18092929, s19010063, 9893336, rs14235986}.
    \item[-] Utilizing open-access SAR datasets for pre-training before fine-tuning \cite{9805965, 10239480}.
    \item[-] Merging optical and SAR imagery to see whether the model performs better at feature extraction, thus leading to improved overall performance \cite{10433638, 9180260, 10497597, 10246308, 11045289}.
\end{itemize}

Transfer learning comes with the challenge of selecting the appropriate model, as numerous architectures are pretrained on large datasets. However, these models may not perform well on SAR ship classification due to the inherent imbalance in SAR data. Additionally, the depth of the networks can lead to varying classification performances. Furthermore, the nature of the data on which the model was pretrained can significantly impact its performance. Therefore, it is crucial to carefully evaluate the pretraining dataset before utilizing any pretrained network.

\subsubsection{Learning Rate management}\label{sec:lrm}
One crucial component of DL is the learning rate, which helps the network to update its weights efficiently. Traditionally, the learning rate is kept fixed during the entire training. However, as a deep neural network progresses through training iterations, it gradually learns information from the data. A constant learning rate in this scenario can slow down training or potentially cause the model to miss the point of lowest error. To address this issue, a strategy was introduced in \cite{9805965} that changes the learning rate based on the number of iterations. This approach reduces the learning rate as training progresses, facilitating convergence towards the optimal weights. This technique not only reduces the training time but can also improve the overall performance of deep neural networks. In tasks like SAR ship classification, where data availability is limited, such dynamic learning rate adjustments can be particularly beneficial for achieving better performances.

\subsubsection{Searched Networks}
The architecture of a DL model is traditionally determined through the researcher's intuition or by replicating existing architectures. However, Neural Architecture Search (NAS) offers a more objective approach to identify the optimal architecture for a specific task. NAS is a search algorithm that explores a range of possible network configurations to identify the one that performs best for the given task. This approach empowers researchers to discover the most suitable architecture for their data, potentially leading to significant advancements in maritime monitoring \cite{app12146866, rs15112904}. While searched networks are an excellent option, they come with a significant computational cost. The more the layers included in the search process, the higher the computational demand. Therefore, it is crucial to carefully consider the computational cost before applying NAS.

\section{Design Considerations}\label{sec6}

\subsection{Model Architectures}
For designing custom CNNs, researchers have proposed shallow DL architectures. These architectures typically consist of 3-5 convolutional layers followed by pooling and activation functions, ultimately connecting to 2-3 fully connected linear layers. NAS can be a valuable tool to identify the optimal architecture for a specific task. In contrast, other proposed architectures leverage established pre-trained models: 
\begin{itemize}
    \item[-] \textbf{AlexNet}~\cite{krizhevsky2012imagenet} is a pioneering deep CNN with 8 layers: 5 convolutional layers followed by 3 fully connected layers. It introduced the use of ReLU activation functions and dropout for regularization.
    \item[-] \textbf{VGGNet}~\cite{simonyan2014very} features a 16-layer version, \mbox{VGGNet-16}, a deep CNN architecture including 13 convolutional layers and 3 fully connected layers. The convolutional layers use small 3x3 receptive fields and are stacked on top of each other to increase the depth of the network. The network also includes five max pooling layers that follow some of the convolutional layers. Its 19-layer version, VGGNet-19, includes three additional convolutional layers. Like \mbox{VGGNet-16}, \mbox{VGGNet-19} uses small 3x3 receptive fields and includes max pooling layers.
    \item[-] \textbf{ResNet}~\cite{he2016deep} features a 50-layer and a 34-layer versions, \mbox{ResNet-50} and \mbox{Resnet-34}, respectively, utilizing residual learning to mitigate the vanishing gradient problem commonly encountered in deep networks. This architecture is composed of convolutional layers organized into residual blocks, each containing shortcut connections that bypass one or more layers. These shortcut connections enable the network to learn identity mappings, thus facilitating the training of much deeper networks by improving gradient flow and convergence.
    \item[-] \textbf{DenseNet-101},~or Dense Convolutional Network \cite{huang2017densely}, comprises 101 layers with a unique dense connectivity pattern. Each layer receives inputs from all preceding layers and passes its own feature maps to all subsequent layers, promoting feature reuse and mitigating the vanishing gradient problem. This dense connectivity reduces the number of parameters and encourages the network to learn more compact and efficient representations.
    \item[-] \textbf{InceptionNet},~also known as GoogLeNet \cite{szegedy2015going}, introduces the inception module, which allows for multi-scale processing by applying parallel convolutional filters of different sizes ($1\times 1$, $3 \times 3$ and $5\times 5$) within the same layer. This architecture comprises 22 layers, with inception modules stacked sequentially to capture diverse spatial features. This design efficiently manages computational resources while enhancing the network's ability to learn complex and varied patterns in the input data.
\end{itemize}
These models are often adapted for SAR data processing through transfer learning approaches. Additionally, generative adversarial networks have also been explored for potential applications in SAR ship classification \cite{9180260, 10246308}.

\subsection{Train-Test Split}
The literature commonly uses an 80-20 split for training and testing, respectively. However, unbalanced data poses a challenge for model learning.  An effective approach, as seen in the literature \cite{rs15082138, rs15112904}, involves using the class with the fewest samples to determine the size of the training set. The remaining data are used for testing. An example of train-test ratio for class imbalanced dataset is visualized in Table~\ref{tab:tt_ratio}. This approach ensures that the model learns balanced information from all classes. However, in cases of extreme imbalance, the training set becomes very limited. Therefore, it is crucial to understand the data complexity before applying such an approach. 
\begin{table}[h]
\centering
\caption{Numbers of ship-type occurrences for imbalanced and balanced train-test splits (example)}
\label{tab:tt_ratio}
\begin{tabular}{|llll|}
\hline
\multicolumn{4}{|l|}{Without considering Class Imbalance}                                                                                                                                                                                    \\ \hline
\multicolumn{1}{|l|}{Class}   & \multicolumn{1}{l|}{\begin{tabular}[c]{@{}l@{}}Total\\ 100\%\end{tabular}} & \multicolumn{1}{l|}{\begin{tabular}[c]{@{}l@{}}Train\\ 80\%\end{tabular}} & \begin{tabular}[c]{@{}l@{}}Test\\ 20\%\end{tabular} \\ \hline
\multicolumn{1}{|l|}{Cargo}   & \multicolumn{1}{l|}{1500}                                                  & \multicolumn{1}{l|}{1200}                                                 & 300                                                 \\ \hline
\multicolumn{1}{|l|}{Tanker}  & \multicolumn{1}{l|}{900}                                                   & \multicolumn{1}{l|}{720}                                                  & 180                                                 \\ \hline
\multicolumn{1}{|l|}{Fishing} & \multicolumn{1}{l|}{600}                                                   & \multicolumn{1}{l|}{480}                                                  & 120                                                 \\ \hline
\multicolumn{4}{|l|}{After considering Class Imbalance}                                                                                                                                                                                      \\ \hline
\multicolumn{1}{|l|}{Class}   & \multicolumn{1}{l|}{\begin{tabular}[c]{@{}l@{}}Total\end{tabular}} & \multicolumn{1}{l|}{\begin{tabular}[c]{@{}l@{}}Train\end{tabular}} & \begin{tabular}[c]{@{}l@{}}Test\end{tabular} \\ \hline
\multicolumn{1}{|l|}{Cargo}   & \multicolumn{1}{l|}{1500}                                                  & \multicolumn{1}{l|}{480}                                                  & 1020                                                \\ \hline
\multicolumn{1}{|l|}{Tanker}  & \multicolumn{1}{l|}{900}                                                   & \multicolumn{1}{l|}{480}                                                  & 420                                                 \\ \hline
\multicolumn{1}{|l|}{Fishing} & \multicolumn{1}{l|}{600}                                                   & \multicolumn{1}{l|}{480}                                                  & 120                                                 \\ \hline
\end{tabular}
\end{table}
\subsection{Hyperparameters}
\subsubsection{\textbf{Epochs}}
In DL, an epoch is a fundamental unit of training, defined as one complete iteration over the entire training set. During each epoch, the neural network processes every sample in the dataset, enabling the model to learn and adjust its parameters—such as weights and biases—based on the error between its predictions and the actual target values. The role of epochs is critical in the iterative optimization process, as each epoch provides an opportunity for the model to refine its internal representations and progressively minimize the loss function. 

The importance of epochs lies in balancing the trade-off between underfitting and overfitting; a sufficient number of epochs ensures that the model captures the underlying patterns in the data, while avoiding excessive training that might lead to overfitting, where the model performs well on the training data but poorly on new, unseen data. Properly tuning the number of epochs, often through techniques such as early stopping, is essential to achieve optimal model performance and generalization. The literature suggests a common range of 50–200 epochs for DL model convergence, but the optimal number ultimately depends on the chosen learning strategy \cite{10457854, 10440349, 10433638, 10246308, 10242360, 10239480, 10189816}.
\subsubsection{\textbf{Batch Size}}
In DL, batch size refers to the number of training samples processed before updating the model parameters. It affects training stability and efficiency: smaller batch sizes offer more frequent updates and can improve generalization, but may introduce noise, while larger batch sizes provide more stable and accurate gradient estimates but require more memory and can slow down the training per epoch. Thus, selecting an appropriate batch size is crucial for balancing training speed, computational resources, and model performance. In the literature, batch sizes of 64, 32 and 16 are commonly used \cite{10487903, 10440349, 10315181, 10242360, 10239480, 10189816, 9782459}.
\subsubsection{\textbf{Image Size}}
Selecting image size plays a crucial role in determining the network computational requirements, memory usage, and ability to capture detailed features. Larger image sizes can provide more detailed spatial information, enhancing the model's ability to detect fine-grained patterns, but they also significantly increase computational load and memory consumption. Conversely, smaller image sizes reduce computational demands and memory usage but may result in the loss of important visual details, potentially degrading the model performance. The literature commonly utilizes image sizes of $224\times 224$, $64\times 64$ and $128\times 128$ \cite{10487903, 10457854, 10315181, 10246308}.

\subsubsection{\textbf{Learning Rate}}
The Learning Rate (see also Section~\ref{sec:lrm}) determines the step size at each iteration while moving towards a minimum of a loss function. It controls the weights of a DL network with respect to the loss gradient. A high learning rate can accelerate training by making large updates to the parameters, but risks overshooting the optimal values, leading to divergence or instability. Conversely, a low learning rate ensures more precise adjustments, promoting stable convergence, but can make the training process exceedingly slow and may get stuck in local minima. Therefore, selecting an appropriate learning rate is critical for achieving efficient and effective training. In the literature, the common learning rate is usually set in the range 0.0001 to 0.001 \cite{10487903, 10457854, 10433638, 10315181, 10246308, 10189816}; moreover, learning rate management can be employed to train the model faster and more precisely \cite{10242360, 10239480}.

\subsection{Number of Classes}
Due to the imbalanced dataset problem in SAR ship data, an increase in the number of classes results in fewer samples per class. For instance, in both the public datasets considered here (FUSARShip and OpenSARShip), over 80\% of the data are concentrated in just five classes on average, with the remaining data sparsely distributed among other classes. This imbalance causes classes with fewer samples to have negligible impact during training, leading the model to overfit towards the more prevalent classes. Consequently, most research focuses on 3-5 classes (see tables-\ref{tab:sar_small_classcount}, \ref{tab:sar_large_classcount}); although a few studies have explored the usage of more than five classes, it is common experience that the model performance significantly deteriorates as the number of classes increases.

\begin{table*}
\centering
\caption{SAR ship-classification results on public datasets with
$\le\!3$ target classes.  Yellow cells mark the best Overall Accuracy (OA) and
F$_1$ within each class-count block.  %
\emph{Method abbreviations}:\;%
CNN—Convolutional Neural Network; %
DA—Domain Adaptation; %
FF—Feature Fusion; %
GAN—Generative Adversarial Network; %
HC—Hand-Crafted features; %
KD—Knowledge Distillation; %
ML—Metric Learning; %
MS—Multi-Scale; %
NAS—Neural Architecture Search; %
PF—Polarization Fusion; %
\emph{Dataset abbreviations}:\;%
CSK—\textit{COSMO-SkyMed}; %
FS—\textit{FUSAR-Ship}; %
GF-3—\textit{Gaofen-3}; %
OS—\textit{OpenSARShip}; %
TX—\textit{TerraSAR-X}.}
\label{tab:sar_small_classcount}
\scriptsize
\begin{adjustbox}{width=\textwidth}
\begin{tabular}{p{6.3cm} p{1.7cm} c c c}
\toprule
\textbf{Method} & \textbf{Dataset} & \textbf{\#Cls.} & \textbf{OA\,[\%]} & \textbf{F$_1$\,[\%]}\\
\midrule
PF (2021)\cite{10027936}                 & GF-3 & 2 & \best{96.55} & \best{93.88} \\
HC FF (2023)\cite{10254064}              & OS   & 2 & 88.87        & 83.49        \\
\midrule
CNN (2018)\cite{s18092929}               & CSK  & 3 & 97.66        & \best{97.80} \\
DA+TL (2019)\cite{s18092929}             & TX   & 3 & \best{98.52} & 97.15        \\
Inc.\ Angle (2019)\cite{8899277}         & CSK  & 3 & 98.13        & --           \\
Upsampling (2019)\cite{8897831}          & OS   & 3 & 81.56        & --           \\
ML (2020)\cite{9199258}                  & OS   & 3 & 83.67        & --           \\
GAN (2020)\cite{9180260}                 & GF-3 & 3 & 95.00        & --           \\
FF (2021)\cite{9553192}                  & OS   & 3 & 77.69        & --           \\
AIS+CNN (2021)\cite{electronics10101169} & OS   & 3 & 93.30        & 93.73        \\
ML (2021)\cite{9146990}                  & OS   & 3 & 88.97        & --           \\
Attn.\ FF (2021)\cite{9564861}           & OS   & 3 & 92.66        & --           \\
FF (2021)\cite{rs13112091}               & OS   & 3 & 78.00        & --           \\
FF (2021)\cite{rs13112091}               & FS   & 3 & 86.86        & --           \\
MS+PF (2021)\cite{9568903}               & OS   & 3 & 79.25        & 77.62        \\
FF (2021)\cite{9688498}                  & OS   & 3 & 77.69        & --           \\
MS+HC FF (2022)\cite{9445223}            & OS   & 3 & 78.15        & 75.04        \\
FF (2022)\cite{9893336}          & TX   & 3 & 88.89        & --           \\
FF (2022)\cite{9893336}          & FS   & 3 & 89.11        & --           \\
PF (2022)\cite{9885840}                  & OS   & 3 & 88.00        & --           \\
PF (2022)\cite{9782459}                  & OS   & 3 & 88.80        & 86.81        \\
NAS (2022)\cite{app12146866}             & OS   & 3 & 88.02        & --           \\
PF+HC (2022)\cite{ZHANG2022108365}       & OS   & 3 & 79.84        & 78.00        \\
DA (2022)\cite{rs14235986}               & TX   & 3 & 87.67        & --           \\
NAS (2023)\cite{rs15112904}              & OS   & 3 & 82.06        & --           \\
PF (2023)\cite{rs15082138}               & OS   & 3 & 81.28        & --           \\
MS+PF (2023)\cite{10149595}              & OS   & 3 & 88.93        & 86.54        \\
Novel CNN (2023)\cite{10242360}          & GF-3 & 3 & 89.79        & 89.71        \\
Attn.\ Dense GAN (2023)\cite{10246308}   & FS   & 3 & 92.00        & --           \\
MS (2023)\cite{10315181}                 & OS   & 3 & 82.23        & 80.91        \\
Dual-Pol (2024)\cite{10457854}           & OS   & 3 & 87.62        & 86.74        \\
DA (2024)\cite{10547709}                 & FS   & 3 & 75.94        & --           \\
Novel Arch. (2024)\cite{10642068}        & OS   & 3 & 81.86        & 80.75        \\
Novel Arch. (2024)\cite{10642068}        & FS   & 3 & 75.18        & 73.05        \\
PF (2024)\cite{10440349}                 & OS   & 3 & 87.13        & 86.27        \\
PF (2024)\cite{rs16183479}               & OS   & 3 & 89.40        & 86.89        \\
FF (2024)\cite{10671463}                 & OS   & 3 & 82.00        & 82.00        \\
FF (2024)\cite{10671463}                 & FS   & 3 & 90.00        & 89.00        \\
MS (2024)\cite{10412205}                 & OS   & 3 & 85.66        & 84.90        \\
MS (2024)\cite{10412205}                 & FS   & 3 & 85.26        & 86.17        \\
B-CNN (2024)\cite{10641086}              & OS   & 3 & 76.30        & 76.30        \\
FM (2024)\cite{10578024}                 & OS   & 3 & 79.02        & --           \\
KD (2024)\cite{10446152}                 & OS   & 3 & 80.03        & --           \\
Imbal. (2024)\cite{10701968}             & FS+OS& 3 & 74.60        & 67.50        \\
CV-DL (2025)\cite{10890957}              & GF-3 & 3 & 72.22        & --           \\
DA (2025)\cite{11045289}                 & FS   & 3 & 95.33        & --           \\
DA (2025)\cite{11045289}                 & TX   & 3 & 89.33        & --           \\
MS FF (2025)\cite{10937719}              & OS   & 3 & 80.09        & 79.27        \\
Super-Res (2025)\cite{11043629}                 & OS   & 3 & 72.94        & 61.53        \\
Super-Res (2025)\cite{11018437}                 & OS   & 3 & 87.68        & --           \\
\bottomrule
\end{tabular}
\end{adjustbox}
\end{table*}

\begin{table*}
\centering
\caption{SAR ship-classification results on public datasets with $>\!3$ target classes.  Abbreviations as in Table~\ref{tab:sar_small_classcount}.}
\label{tab:sar_large_classcount}
\scriptsize
\begin{adjustbox}{width=\textwidth}
\begin{tabular}{p{6.3cm} p{1.7cm} c c c}
\toprule
\textbf{Method} & \textbf{Dataset} & \textbf{\#Cls.} & \textbf{OA.\,[\%]} & \textbf{F$_1$\,[\%]}\\
\midrule
CNN\,+\,Spatial Attention (2018)\cite{s18093039}      & OS   & 4 & 84.00 & -- \\
CNN (2019)\cite{10.1007/978-3-030-01054-6_2}          & P2   & 4 & 65.70 & -- \\
Deep FF (2019)\cite{9172933}                          & OS   & 4 & 87.04 & -- \\
FF (2021)\cite{RELEKAR20214594}                       & OS   & 4 & 82.35 & 82.19 \\
DA (2022)\cite{rs14235986}                            & FS   & 4 & 82.33 & -- \\
NAS (2023)\cite{rs15112904}                           & FS   & 4 & 63.90 & -- \\
HC (2024)\cite{10487903}                              & OS   & 4 & \best{87.50} & \best{86.00} \\
LSTM\,+\,CNN (2024)\cite{s24247954}                   & FS   & 4 & 65.58 & -- \\
Complex-Valued DL (2025)\cite{10890957}               & GF-3 & 4 & 69.01 & -- \\
\midrule
CNN (2018)\cite{8113469}                              & TX   & 5 & --    & 94.00 \\
Metric Learning (2021)\cite{9146990}                  & OS   & 5 & 68.16 & -- \\
PF (2022)\cite{9782459}                               & OS   & 5 & 66.90 & 57.54 \\
Novel CNN (2023)\cite{rs15112917}                     & FS   & 5 & \best{98.71} & \best{97.69} \\
MS\,+\,PF (2023)\cite{10149595}                       & OS   & 5 & 67.03 & 57.89 \\
Data Augmentation (2024)\cite{rs16071299}             & FS   & 5 & 72.19 & 71.97 \\
PF (2024)\cite{rs16183479}                            & OS   & 5 & 67.83 & 58.01 \\
DA (2024)\cite{10497597}                              & FS   & 5 & 91.22 & -- \\
MS (2024)\cite{10412205}                              & FS   & 5 & 84.24 & 84.36 \\
Metric Learning (2024)\cite{10433638}                 & FS   & 5 & 78.57 & -- \\
\midrule
MS\,+\,PF (2021)\cite{9568903}                        & OS   & 6 & 56.66 & 51.38 \\
Sampling (2021)\cite{9390936}                         & OS   & 6 & 80.58 & -- \\
NAS (2022)\cite{ZHANG2022108365}                      & OS   & 6 & 56.73 & -- \\
PF\,+\,HC (2022)\cite{ZHANG2022108365}                & OS   & 6 & 56.83 & 51.82 \\
PF (2023)\cite{rs15082138}                            & OS   & 6 & 58.68 & -- \\
MS (2023)\cite{10147843}                              & OS   & 6 & \best{81.83} & -- \\
MS (2023)\cite{10315181}                              & OS   & 6 & 60.76 & 59.89 \\
Dual-Pol (2024)\cite{10457854}                        & OS   & 6 & 64.26 & 58.44 \\
PF (2024)\cite{10440349}                              & OS   & 6 & 65.97 & 58.63 \\
MS\,FF (2025)\cite{10937719}                          & OS   & 6 & 60.63 & 58.29 \\
Super-Res (2025)\cite{11018437}                       & OS   & 6 & 69.71 & -- \\
Super-Res (2025)\cite{awais2025classification}                       & OS   & 6 & -- & \best{65.40} \\
\midrule
FF (2021)\cite{rs13112091}                            & FS   & 7 & 86.86 & -- \\
MS\,+\,HC\,FF (2022)\cite{9445223}                    & FS   & 7 & 86.69 & 86.58 \\
HC\,FF (2023)\cite{10189816}                          & FS   & 7 & 87.23 & 86.69 \\
MS (2023)\cite{10315181}                              & FS   & 7 & \best{90.08} & \best{90.12} \\
Feature Manip. (2024)\cite{10578024}                  & FS   & 7 & 87.59 & -- \\
Knowledge Distill. (2024)\cite{10446152}              & FS   & 7 & 86.07 & -- \\
HC (2024)\cite{10446741}                              & FS   & 7 & 87.94 & -- \\
\midrule
Deep FF (2025)\cite{10908205}                         & FS   & 8  & 76.12 & \best{56.67} \\
Curriculum Learning (2025)\cite{Awais_2025_WACV}      & FS   & 9  & 68.34 & \best{55.49} \\
Over-Sampling (2025)\cite{awais2025feature}                       & FS   & 9 & -- & 52.87 \\
Side-Lobe Elimin. (2021)\cite{9455304}                & FS   & 10 & \best{98.40} & 97.96 \\
Siamese Net (2021)\cite{9512781}                      & OS   & 16 & --    & \best{97.97} \\
\bottomrule
\end{tabular}
\end{adjustbox}
\end{table*}

\begin{table*}[ht]
\centering
\caption{Challenges in SAR ship classification and practical remedies.}
\label{tab:map}
\begin{tabular}{p{3.4cm} p{10.4cm}}
\toprule
Challenge & Remedies / notes \\
\midrule
Data scarcity & High-quality open SAR curation; AIS--SAR mapping for labels; strong augmentations; transfer learning. \\
Class imbalance & Rebalanced splits; focal/Class Balanced loss; mixup/cutmix/feature-space-oversampling; report F1 alongside OA, curriculum learning. \\
Domain shift (sensor/scene) & Normalization; domain adaptation; test-time adaptation; confidence calibration. \\
Small targets / clutter & Multi-scale features; higher input resolution; denoising; polarization/feature fusion, super resolution. \\
Reproducibility & Report splits/seeds/hyperparams; share CSV of results. \\
\bottomrule
\end{tabular}
\end{table*}

\section{Outlook}\label{sec8}

This survey provides a comprehensive review of various techniques employed to train DL models for ship classification using SAR imagery. Through a systematic literature review, we assessed the content of 187  papers, rigorously narrowing down our selection to the most pertinent 74 studies. Their references are collected in Table~\ref{tab:paper_timeline} and their features are summarized in Section~\ref{sec4}, offering an overview of their key elements and contributions to the field. 

The literature reveals a broad range of DL techniques employed for SAR ship classification, as illustrated in Figure~\ref{fig:taxanomy}. These techniques include 
CNNs, in the earliest proposals, 
and state-of-the-art architectures such as ResNet, VGG, DenseNet etc., each one tailored to leverage the unique attributes of SAR images.

Our analysis indicates that the choice of the DL technique significantly impacts classification performance. For instance, shallow CNNs have been particularly effective to learn spatial hierarchies from SAR images, as the data are scarce.


The FUSARShip and OpenSARShip datasets, along with other distinct data sources, are extensively used in the literature. These datasets are crucial for training and evaluating DL models. However, the limited availability of SAR images poses a significant challenge.
To address this, data augmentation techniques have been proposed and widely adopted. These techniques artificially increase the diversity and quantity of training data, which is essential for improving the performance and robustness of DL models. Fine-tuning pre-trained DL models on SAR-specific datasets has emerged as a successful strategy. This approach leverages the knowledge from pre-trained models and adapts it to the specific characteristics of SAR data, resulting in improved classification accuracy.

The issue of data imbalance has been addressed in the literature through two main solutions: adjusting train-test datasets ratios and employing tailored loss functions. Additionally, some researchers propose the use of architecture search techniques like NAS for selecting the optimal DL architecture. This survey also highlights the significance of various training techniques, including feature fusion, polarization fusion, metric learning, multi-scale/pyramid learning, and learning rate management.

The findings of our survey underscore the importance of data quality and the choice of techniques in SAR ship classification. The integration of handcrafted features, effective data augmentation, and fine-tuning of pre-trained models are all critical factors contributing to the success of DL models in this domain. 
Building on the findings of this survey and acknowledging the limitations of current research, several exciting avenues for future work in SAR ship classification using DL are identified and described in the following subsections.

\subsection{Addressing Data scarcity}
The data used by researchers for DL-based SAR ship classification is seldom made public due to privacy or digital rights reasons. Future research should focus on developing methodologies to publicize the data without such constraints. In addition, creating high-quality datasets from open-access SAR imagery and establishing methodologies for reliable AIS–SAR mapping can play a key role in alleviating the problem of data scarcity. Developing more robust data augmentation techniques can further contribute to the diversity of SAR datasets \cite{rs16071299}. Finally, collaborative efforts aimed at standardizing and sharing datasets would significantly benefit the research community.
\subsection{Novel DL architectures}
The development of DL architectures specifically tailored for SAR ship classification holds significant promise for advancing the research in this field. This includes exploring and optimizing various DL architecture components, such as loss functions, learning rates, activation functions, and even utilizing techniques like architecture search and fine-tuning. Furthermore, the continuous evolution of DL architectures presents an exciting opportunity to investigate the efficacy of novel models in SAR ship classification tasks. Recent advancements in areas such as transformers and generative models (e.g., TransGAN \cite{jiang2021transgan}) are particularly promising for tasks such as noise reduction and data augmentation, which can be crucial for improving the performance of DL models in SAR ship classification.

\subsection{Handcrafted and Deep Features Integration}
DL models have achieved remarkable success in feature learning for various tasks. However, handcrafted features, meticulously designed to capture specific characteristics of SAR images, remain a valuable tool. Further research is warranted to explore how these can be effectively integrated with DL models for SAR image classification. This combined approach has the potential to leverage the strengths of both machine learning and DL methods, potentially leading to significant performance improvements. Moreover, leveraging handcrafted features and intrinsic SAR properties may provide promising pathways for the development of next-generation models, including foundation models and vision–language architectures \cite{xiao2025foundation}.
\subsection{Standardized Performance Metrics}
The field of SAR ship classification currently lacks a standardized set of performance metrics, hindering comprehensive comparisons between different studies. Utilizing metrics like the F1 score can partially address this issue, but future research should focus on establishing a common set of metrics specifically tailored to the nuances of SAR ship classification tasks. This would enable more robust evaluations and facilitate meaningful comparisons between different DL models and training strategies.

Furthermore, inconsistencies in data utilization across studies pose another challenge to comparison. Even when researchers employ the same datasets, variations in sample selection can significantly impact the results. Moreover, among the researchers in SAR ship classification, code sharing is not common; if we want to progress further and faster, code sharing should be adopted as a practice. Standardized metrics and data utilization practices are crucial for achieving truly comparable research, ultimately accelerating the advancements in the field.
\subsection{Interpretability of DL models}
DL models, while demonstrably powerful, are often characterized as ``black boxes'' due to their complex internal workings. This lack of interpretability can hinder our understanding of the rationale behind their classification decisions. Fortunately, various interpretability techniques, such as Gradient-weighted Class Activation Mapping (Grad-CAM, \cite{selvaraju2017grad}), SHapley Additive exPlanations (SHAP, \cite{lundberg2017unified}) and Local Interpretable Model-Agnostic Explanations (LIME, \cite{ribeiro2016should}) can be employed to shed light on the internal decision-making processes of these models.

While interpretability has not always been a central focus in SAR ship classification research, it presents a valuable avenue for future exploration. By leveraging interpretability techniques, researchers can gain deeper insights into the features most critical for the model's decisions. This enhanced understanding can lead to more informed analysis, improved trust in model predictions, and even to the identification of potential biases within the model.
\subsection{Interdisciplinary Collaborations}
Significant advancements in SAR ship classification using DL can be achieved by fostering interdisciplinary collaborations between researchers in remote sensing, computer vision, and maritime studies. This collaborative approach has the potential to not only generate innovative solutions for the challenges faced in SAR ship classification, but also contribute to environmental protection efforts by facilitating the detection of illegal maritime activities.

By focusing on these directions, future research can build on our findings and continue to advance the field of SAR ship classification using DL, ultimately improving the accuracy and efficiency of maritime surveillance and related applications, we also present common challenges and remedies to those challenges in table-\ref{tab:map}.

\section{Conclusion}\label{sec9}

This survey has systematically reviewed the diverse techniques employed to train DL models for SAR ship classification. We identified critical trends and challenges in this domain, emphasizing the importance of integrating handcrafted features, utilizing public datasets, data augmentation, fine-tuning, explainability, and interdisciplinary collaborations for enhancing DL model performance.

We established a survey methodology with clearly defined research questions. Our literature search yielded a comprehensive dataset of relevant papers, which were then categorized into a first-of-its-kind taxonomy encompassing DL techniques, handcrafted feature use, SAR attribute utilization, and the impact of fine-tuning.

In particular, Section~\ref{sec4} provides detailed explanations of the included papers, categorized into the four sections (discussing DL techniques, impact of handcrafted features, use of SAR properties and fine-tuning). Section~\ref{sec5} presents the methodologies employed in SAR ship classification.
Finally, Section~\ref{sec8} outlines potential avenues for addressing current challenges, such as data scarcity, novel DL architectures, interpretability, performance metrics, feature integration, and interdisciplinary collaborations.

Advancements in DL techniques for SAR ship classification hold significant promise for improving maritime surveillance and related applications. Continued exploration and innovation in this field can lead to more accurate and efficient ship classification systems, ultimately enhancing the capabilities of maritime operations.









\section*{Acknowledgments}
This work was supported by National Recovery and Resilience Plan (NRRP), Mission 4 Component 2 Investment 1.4 - Call for tender No. 3138 of 16 December 2021, rectified by Decree n.3175 of 18 December 2021  of Italian Ministry of University and Research funded by the European Union – NextGenerationEU. Award Number: Project code CN\_00000033, Concession Decree No. 1034  of 17 June 2022 adopted by the Italian Ministry of University and Research,  CUP D33C22000960007, Project title “National Biodiversity Future Center - NBFC”.



\bibliography{biblio.bib}

\begin{thebibliography}{100}
\providecommand{\url}[1]{#1}
\csname url@samestyle\endcsname
\providecommand{\newblock}{\relax}
\providecommand{\bibinfo}[2]{#2}
\providecommand{\BIBentrySTDinterwordspacing}{\spaceskip=0pt\relax}
\providecommand{\BIBentryALTinterwordstretchfactor}{4}
\providecommand{\BIBentryALTinterwordspacing}{\spaceskip=\fontdimen2\font plus
\BIBentryALTinterwordstretchfactor\fontdimen3\font minus \fontdimen4\font\relax}
\providecommand{\BIBforeignlanguage}[2]{{%
\expandafter\ifx\csname l@#1\endcsname\relax
\typeout{** WARNING: IEEEtran.bst: No hyphenation pattern has been}%
\typeout{** loaded for the language `#1'. Using the pattern for}%
\typeout{** the default language instead.}%
\else
\language=\csname l@#1\endcsname
\fi
#2}}
\providecommand{\BIBdecl}{\relax}
\BIBdecl

\bibitem{espedal1999satellite}
H.~Espedal, ``Satellite \mbox{SAR} oil spill detection using wind history information,'' \emph{International Journal of Remote Sensing}, vol.~20, no.~1, pp. 49--65, 1999.

\bibitem{salberg2014oil}
A.-B. Salberg, {\O}.~Rudjord, and A.~H.~S. Solberg, ``Oil spill detection in hybrid-polarimetric \mbox{SAR} images,'' \emph{IEEE Transactions on Geoscience and Remote Sensing}, vol.~52, no.~10, pp. 6521--6533, 2014.

\bibitem{topouzelis2008oil}
K.~N. Topouzelis, ``Oil spill detection by \mbox{SAR} images: Dark formation detection, feature extraction and classification algorithms,'' \emph{Sensors}, vol.~8, no.~10, pp. 6642--6659, 2008.

\bibitem{li2022deep}
J.~Li, C.~Xu, H.~Su, L.~Gao, and T.~Wang, ``Deep learning for \mbox{SAR} ship detection: Past, present and future,'' \emph{Remote Sensing}, vol.~14, no.~11, p. 2712, 2022.

\bibitem{yasir2023ship}
M.~Yasir, W.~Jianhua, X.~Mingming, S.~Hui, Z.~Zhe, L.~Shanwei, A.~T.~I. Colak, and M.~S. Hossain, ``Ship detection based on deep learning using \mbox{SAR} imagery: a systematic literature review,'' \emph{Soft Computing}, vol.~27, no.~1, pp. 63--84, 2023.

\bibitem{zhang2021sar}
T.~Zhang, X.~Zhang, J.~Li, X.~Xu, B.~Wang, X.~Zhan, Y.~Xu, X.~Ke, T.~Zeng, H.~Su \emph{et~al.}, ``\mbox{SAR} ship detection dataset (\mbox{SSDD}): Official release and comprehensive data analysis,'' \emph{Remote Sensing}, vol.~13, no.~18, p. 3690, 2021.

\bibitem{zilman2004speed}
G.~Zilman, A.~Zapolski, and M.~Marom, ``The speed and beam of a ship from its wake's \mbox{SAR} images,'' \emph{IEEE Transactions on Geoscience and Remote Sensing}, vol.~42, no.~10, pp. 2335--2343, 2004.

\bibitem{kang2019ship}
K.-m. Kang and D.-j. Kim, ``Ship velocity estimation from ship wakes detected using convolutional neural networks,'' \emph{IEEE Journal of Selected Topics in Applied Earth Observations and Remote Sensing}, vol.~12, no.~11, pp. 4379--4388, 2019.

\bibitem{graziano2019integration}
M.~D. Graziano, A.~Renga, and A.~Moccia, ``Integration of automatic identification system (\mbox{AIS}) data and single-channel synthetic aperture radar (\mbox{SAR}) images by \mbox{SAR}-based ship velocity estimation for maritime situational awareness,'' \emph{Remote Sensing}, vol.~11, no.~19, p. 2196, 2019.

\bibitem{heiselberg2023ship}
P.~Heiselberg, K.~S{\o}rensen, and H.~Heiselberg, ``Ship velocity estimation in \mbox{SAR} images using multitask deep learning,'' \emph{Remote Sensing of Environment}, vol. 288, p. 113492, 2023.

\bibitem{xing2013ship}
X.~Xing, K.~Ji, H.~Zou, W.~Chen, and J.~Sun, ``Ship classification in \mbox{TerraSAR-X} images with feature space based sparse representation,'' \emph{IEEE Geoscience and Remote Sensing Letters}, vol.~10, no.~6, pp. 1562--1566, 2013.

\bibitem{jiang2016ship}
M.~Jiang, X.~Yang, Z.~Dong, S.~Fang, and J.~Meng, ``Ship classification based on superstructure scattering features in \mbox{SAR} images,'' \emph{IEEE Geoscience and Remote Sensing Letters}, vol.~13, no.~5, pp. 616--620, 2016.

\bibitem{lang2015ship}
H.~Lang, J.~Zhang, X.~Zhang, and J.~Meng, ``Ship classification in \mbox{SAR} image by joint feature and classifier selection,'' \emph{IEEE Geoscience and Remote Sensing Letters}, vol.~13, no.~2, pp. 212--216, 2015.

\bibitem{wu2017novel}
J.~Wu, Y.~Zhu, Z.~Wang, Z.~Song, X.~Liu, W.~Wang, Z.~Zhang, Y.~Yu, Z.~Xu, T.~Zhang \emph{et~al.}, ``A novel ship classification approach for high resolution \mbox{SAR} images based on the \mbox{BDA-KELM} classification model,'' \emph{International Journal of Remote Sensing}, vol.~38, no.~23, pp. 6457--6476, 2017.

\bibitem{galdelli2020integrating}
A.~Galdelli, A.~Mancini, C.~Ferr{\`a}, and A.~N. Tassetti, ``Integrating \mbox{AIS} and \mbox{SAR} to monitor fisheries: A pilot study in the \mbox{Adriatic Sea},'' in \emph{Proceedings of the 2020 IMEKO TC-19 International Workshop on Metrology for the Sea, Naples, Italy}, 2020, pp. 5--7.

\bibitem{chaturvedi2012ship}
S.~K. Chaturvedi, C.-S. Yang, K.~Ouchi, and P.~Shanmugam, ``Ship recognition by integration of \mbox{SAR} and \mbox{AIS},'' \emph{The Journal of Navigation}, vol.~65, no.~2, pp. 323--337, 2012.

\bibitem{hou2020fusar}
X.~Hou, W.~Ao, Q.~Song, J.~Lai, H.~Wang, and F.~Xu, ``\mbox{FUSAR-Ship}: Building a high-resolution \mbox{SAR-AIS} matchup dataset of \mbox{Gaofen-3} for ship detection and recognition,'' \emph{Science China Information Sciences}, vol.~63, pp. 1--19, 2020.

\bibitem{succi2023meta}
G.~Succi, ``A meta-analytical comparison of naive \mbox{Bayes} and random forest for software defect prediction,'' in \emph{Intelligent Systems Design and Applications: 22nd International Conference on Intelligent Systems Design and Applications (ISDA 2022) Held December 12-14, 2022-Volume 3}, vol. 716.\hskip 1em plus 0.5em minus 0.4em\relax Springer Nature, 2023, p. 139.

\bibitem{petticrew2008systematic}
M.~Petticrew and H.~Roberts, \emph{Systematic reviews in the social sciences: A practical guide}.\hskip 1em plus 0.5em minus 0.4em\relax John Wiley \& Sons, 2008.

\bibitem{8127014}
C.~Wang, H.~Zhang, F.~Wu, B.~Zhang, and S.~Tian, ``Ship classification with deep learning using \mbox{COSMO-SkyMed} \mbox{SAR} data,'' in \emph{2017 IEEE International Geoscience and Remote Sensing Symposium (IGARSS)}, 2017, pp. 558--561.

\bibitem{8113469}
C.~Bentes, D.~Velotto, and B.~Tings, ``Ship classification in \mbox{TerraSAR-X} images with convolutional neural networks,'' \emph{IEEE Journal of Oceanic Engineering}, vol.~43, no.~1, pp. 258--266, 2018.

\bibitem{s18092929}
\BIBentryALTinterwordspacing
Y.~Wang, C.~Wang, and H.~Zhang, ``Ship classification in high-resolution \mbox{SAR} images using deep learning of small datasets,'' \emph{Sensors}, vol.~18, no.~9, 2018. [Online]. Available: \url{https://www.mdpi.com/1424-8220/18/9/2929}
\BIBentrySTDinterwordspacing

\bibitem{s18093039}
\BIBentryALTinterwordspacing
J.~Shao, C.~Qu, J.~Li, and S.~Peng, ``A lightweight convolutional neural network based on visual attention for \mbox{SAR} image target classification,'' \emph{Sensors}, vol.~18, no.~9, 2018. [Online]. Available: \url{https://www.mdpi.com/1424-8220/18/9/3039}
\BIBentrySTDinterwordspacing

\bibitem{s19010063}
\BIBentryALTinterwordspacing
C.~Lu and W.~Li, ``Ship classification in high-resolution \mbox{SAR} images via transfer learning with small training dataset,'' \emph{Sensors}, vol.~19, no.~1, 2019. [Online]. Available: \url{https://www.mdpi.com/1424-8220/19/1/63}
\BIBentrySTDinterwordspacing

\bibitem{8899277}
Y.~Dong, C.~Wang, H.~Zhang, Y.~Wang, and B.~Zhang, ``Impact analysis of incident angle factor on high-resolution \mbox{SAR} image ship classification based on deep learning,'' in \emph{IGARSS 2019 - 2019 IEEE International Geoscience and Remote Sensing Symposium}, 2019, pp. 1358--1361.

\bibitem{8897831}
Y.~Wu, Y.~Yuan, J.~Guan, L.~Yin, J.~Chen, G.~Zhang, and P.~Feng, ``Joint convolutional neural network for small-scale ship classification in \mbox{SAR} images,'' in \emph{IGARSS 2019 - 2019 IEEE International Geoscience and Remote Sensing Symposium}, 2019, pp. 2619--2622.

\bibitem{10.1007/978-3-030-01054-6_2}
S.~Hashimoto, Y.~Sugimoto, K.~Hamamoto, and N.~Ishihama, ``Ship classification from \mbox{SAR} images based on deep learning,'' in \emph{Intelligent Systems and Applications}, K.~Arai, S.~Kapoor, and R.~Bhatia, Eds.\hskip 1em plus 0.5em minus 0.4em\relax Cham: Springer International Publishing, 2019, pp. 18--34.

\bibitem{9172933}
Y.~Xi, G.~Xiong, and W.~Yu, ``Feature-loss double fusion siamese network for dual-polarized \mbox{SAR} ship classification,'' in \emph{2019 IEEE International Conference on Signal, Information and Data Processing (ICSIDP)}, 2019, pp. 1--5.

\bibitem{Zhu_2020}
\BIBentryALTinterwordspacing
H.~Zhu, N.~Lin, and D.~Leung, ``Ship classification from \mbox{SAR} images based on sequence input of deep neural network,'' \emph{Journal of Physics: Conference Series}, vol. 1549, no.~5, p. 052042, jun 2020. [Online]. Available: \url{https://dx.doi.org/10.1088/1742-6596/1549/5/052042}
\BIBentrySTDinterwordspacing

\bibitem{9199258}
Y.~Li, X.~Li, Q.~Sun, and Q.~Dong, ``\mbox{SAR} image classification using \mbox{CNN} embeddings and metric learning,'' \emph{IEEE Geoscience and Remote Sensing Letters}, vol.~19, pp. 1--5, 2020.

\bibitem{9553192}
T.~Zhang, X.~Zhang, J.~Shi, and S.~Wei, ``A \mbox{HOG} feature fusion method to improve \mbox{CNN}-based \mbox{SAR} ship classification accuracy,'' in \emph{2021 IEEE International Geoscience and Remote Sensing Symposium IGARSS}, 2021, pp. 5311--5314.

\bibitem{electronics10101169}
\BIBentryALTinterwordspacing
H.-K. Jeon and C.-S. Yang, ``Enhancement of ship type classification from a combination of \mbox{CNN} and \mbox{KNN},'' \emph{Electronics}, vol.~10, no.~10, 2021. [Online]. Available: \url{https://www.mdpi.com/2079-9292/10/10/1169}
\BIBentrySTDinterwordspacing

\bibitem{9553116}
X.~Xu, X.~Zhang, and T.~Zhang, ``Multi-scale \mbox{SAR} ship classification with convolutional neural network,'' in \emph{2021 IEEE International Geoscience and Remote Sensing Symposium IGARSS}, 2021, pp. 4284--4287.

\bibitem{9146990}
J.~He, Y.~Wang, and H.~Liu, ``Ship classification in medium-resolution \mbox{SAR} images via densely connected triplet \mbox{CNNs} integrating \mbox{Fisher} discrimination regularized metric learning,'' \emph{IEEE Transactions on Geoscience and Remote Sensing}, vol.~59, no.~4, pp. 3022--3039, 2021.

\bibitem{9554116}
Y.~Zhang, Q.~Hua, Y.~Jiang, H.~Li, and D.~Xu, ``\mbox{CV-MotionNet}: Complex-valued convolutional neural network for \mbox{SAR} moving ship targets classification,'' in \emph{2021 IEEE International Geoscience and Remote Sensing Symposium IGARSS}, 2021, pp. 4280--4283.

\bibitem{9564861}
G.~Zhou, G.~Zhang, Z.~Fang, and Q.~Dai, ``A multiscale dual-attention based convolutional neural network for ship classification in \mbox{SAR} image,'' in \emph{2021 IEEE International Conference on Signal Processing, Communications and Computing (\mbox{ICSPCC})}, 2021, pp. 1--5.

\bibitem{rs13112091}
\BIBentryALTinterwordspacing
T.~Zhang and X.~Zhang, ``Injection of traditional hand-crafted features into modern \mbox{CNN}-based models for \mbox{SAR} ship classification: What, why, where, and how,'' \emph{Remote Sensing}, vol.~13, no.~11, 2021. [Online]. Available: \url{https://www.mdpi.com/2072-4292/13/11/2091}
\BIBentrySTDinterwordspacing

\bibitem{9455304}
H.~Zhu, ``Ship classification based on sidelobe elimination of \mbox{SAR} images supervised by visual model,'' in \emph{2021 IEEE Radar Conference (RadarConf21)}, 2021, pp. 1--6.

\bibitem{RELEKAR20214594}
\BIBentryALTinterwordspacing
H.~Relekar and P.~Shanmugam, ``Transfer learning based ship classification in \mbox{Sentinel-1} images incorporating scale variant features,'' \emph{Advances in Space Research}, vol.~68, no.~11, pp. 4594--4615, 2021. [Online]. Available: \url{https://www.sciencedirect.com/science/article/pii/S027311772100692X}
\BIBentrySTDinterwordspacing

\bibitem{9568903}
T.~Zhang and X.~Zhang, ``Squeeze-and-excitation laplacian pyramid network with dual-polarization feature fusion for ship classification in \mbox{SAR} images,'' \emph{IEEE Geoscience and Remote Sensing Letters}, vol.~19, pp. 1--5, 2021.

\bibitem{10027936}
Z.~Sun, B.~Xiong, Y.~Lei, X.~Leng, and K.~Ji, ``Ship classification in high-resolution \mbox{SAR} images based on \mbox{CNN} regional feature fusion,'' in \emph{2021 \mbox{CIE} International Conference on Radar (Radar)}, 2021, pp. 1445--1449.

\bibitem{9170851}
C.~Wang, J.~Shi, Y.~Zhou, X.~Yang, Z.~Zhou, S.~Wei, and X.~Zhang, ``Semisupervised learning-based \mbox{SAR} \mbox{ATR} via self-consistent augmentation,'' \emph{IEEE Transactions on Geoscience and Remote Sensing}, vol.~59, no.~6, pp. 4862--4873, 2021.

\bibitem{9390936}
Y.~Zhang, Z.~Lei, L.~Zhuang, and H.~Yu, ``A \mbox{CNN} based method to solve class imbalance problem in \mbox{SAR} image ship target recognition,'' in \emph{2021 IEEE 5th Advanced Information Technology, Electronic and Automation Control Conference (IAEAC)}, vol.~5, 2021, pp. 229--233.

\bibitem{9688498}
T.~Zhang and X.~Zhang, ``Integrate traditional hand-crafted features into modern \mbox{CNN}-based models to further improve \mbox{SAR} ship classification accuracy,'' in \emph{2021 7th Asia-Pacific Conference on Synthetic Aperture Radar (APSAR)}, 2021, pp. 1--6.

\bibitem{9387401}
L.~Zeng, Q.~Zhu, D.~Lu, T.~Zhang, H.~Wang, J.~Yin, and J.~Yang, ``Dual-polarized \mbox{SAR} ship grained classification based on \mbox{CNN} with hybrid channel feature loss,'' \emph{IEEE Geoscience and Remote Sensing Letters}, vol.~19, pp. 1--5, 2022.

\bibitem{9445223}
T.~Zhang, X.~Zhang, X.~Ke, C.~Liu, X.~Xu, X.~Zhan, C.~Wang, I.~Ahmad, Y.~Zhou, D.~Pan, J.~Li, H.~Su, J.~Shi, and S.~Wei, ``\mbox{HOG-ShipCLSNet}: A novel deep learning network with \mbox{HOG} feature fusion for \mbox{SAR} ship classification,'' \emph{IEEE Transactions on Geoscience and Remote Sensing}, vol.~60, pp. 1--22, 2022.

\bibitem{9893336}
S.~Zhao and H.~Lang, ``Improving deep subdomain adaptation by dual-branch network embedding attention module for \mbox{SAR} ship classification,'' \emph{IEEE Journal of Selected Topics in Applied Earth Observations and Remote Sensing}, vol.~15, pp. 8038--8048, 2022.

\bibitem{9180260}
L.~Li, C.~Wang, H.~Zhang, and B.~Zhang, ``\mbox{SAR} image ship object generation and classification with improved residual conditional generative adversarial network,'' \emph{IEEE Geoscience and Remote Sensing Letters}, vol.~19, pp. 1--5, 2022.

\bibitem{9885840}
J.~He, W.~Chang, F.~Wang, Q.~Wang, Y.~Li, and Y.~Gan, ``Polarization matters: On bilinear convolutional neural networks for ship classification from synthetic aperture radar images,'' in \emph{2022 4th International Conference on Natural Language Processing (ICNLP)}, 2022, pp. 315--319.

\bibitem{9805965}
A.~Toumi, J.-C. Cexus, and A.~Khenchaf, ``A proposal learning strategy on \mbox{CNN} architectures for targets classification,'' in \emph{2022 6th International Conference on Advanced Technologies for Signal and Image Processing (ATSIP)}, 2022, pp. 1--6.

\bibitem{9512781}
J.~A. Raj, S.~M. Idicula, and B.~Paul, ``One-shot learning-based \mbox{SAR} ship classification using new hybrid siamese network,'' \emph{IEEE Geoscience and Remote Sensing Letters}, vol.~19, pp. 1--5, 2022.

\bibitem{9782459}
J.~He, W.~Chang, F.~Wang, Y.~Liu, Y.~Wang, H.~Liu, Y.~Li, and L.~Liu, ``Group bilinear \mbox{CNNs} for dual-polarized \mbox{SAR} ship classification,'' \emph{IEEE Geoscience and Remote Sensing Letters}, vol.~19, pp. 1--5, 2022.

\bibitem{app12146866}
\BIBentryALTinterwordspacing
H.~Zhu, S.~Guo, W.~Sheng, and L.~Xiao, ``\mbox{SBNN}: A searched binary neural network for \mbox{SAR} ship classification,'' \emph{Applied Sciences}, vol.~12, no.~14, 2022. [Online]. Available: \url{https://www.mdpi.com/2076-3417/12/14/6866}
\BIBentrySTDinterwordspacing

\bibitem{ZHANG2022108365}
\BIBentryALTinterwordspacing
T.~Zhang and X.~Zhang, ``A polarization fusion network with geometric feature embedding for \mbox{SAR} ship classification,'' \emph{Pattern Recognition}, vol. 123, p. 108365, 2022. [Online]. Available: \url{https://www.sciencedirect.com/science/article/pii/S0031320321005458}
\BIBentrySTDinterwordspacing

\bibitem{rs14235986}
\BIBentryALTinterwordspacing
H.~Lang, R.~Wang, S.~Zheng, S.~Wu, and J.~Li, ``Ship classification in \mbox{SAR} imagery by shallow \mbox{CNN} pre-trained on task-specific dataset with feature refinement,'' \emph{Remote Sensing}, vol.~14, no.~23, 2022. [Online]. Available: \url{https://www.mdpi.com/2072-4292/14/23/5986}
\BIBentrySTDinterwordspacing

\bibitem{rs15112904}
\BIBentryALTinterwordspacing
H.~Zhu, S.~Guo, W.~Sheng, and L.~Xiao, ``\mbox{SCM}: A searched convolutional metaformer for \mbox{SAR} ship classification,'' \emph{Remote Sensing}, vol.~15, no.~11, 2023. [Online]. Available: \url{https://www.mdpi.com/2072-4292/15/11/2904}
\BIBentrySTDinterwordspacing

\bibitem{rs15112917}
\BIBentryALTinterwordspacing
H.~Guo and L.~Ren, ``A marine small-targets classification algorithm based on improved convolutional neural networks,'' \emph{Remote Sensing}, vol.~15, no.~11, 2023. [Online]. Available: \url{https://www.mdpi.com/2072-4292/15/11/2917}
\BIBentrySTDinterwordspacing

\bibitem{rs15082138}
\BIBentryALTinterwordspacing
Z.~Shao, T.~Zhang, and X.~Ke, ``A dual-polarization information-guided network for \mbox{SAR} ship classification,'' \emph{Remote Sensing}, vol.~15, no.~8, 2023. [Online]. Available: \url{https://www.mdpi.com/2072-4292/15/8/2138}
\BIBentrySTDinterwordspacing

\bibitem{10149595}
J.~He, W.~Chang, F.~Wang, Y.~Liu, C.~Sun, and Y.~Li, ``Multi-scale dense networks for ship classification using dual-polarization \mbox{SAR} images,'' in \emph{2023 IEEE Radar Conference (RadarConf23)}, 2023, pp. 1--6.

\bibitem{10147843}
M.~Yang, X.~Bai, L.~Wang, and F.~Zhou, ``\mbox{HENC}: Hierarchical embedding network with center calibration for few-shot fine-grained \mbox{SAR} target classification,'' \emph{IEEE Transactions on Image Processing}, vol.~32, pp. 3324--3337, 2023.

\bibitem{10433638}
J.~Xu and H.~Lang, ``A unified multiple proxy deep metric learning framework embedded with distribution optimization for fine-grained ship classification in remote sensing images,'' \emph{IEEE Journal of Selected Topics in Applied Earth Observations and Remote Sensing}, vol.~17, pp. 5604--5620, 2024.

\bibitem{10283157}
Y.~Qi, L.~Wang, C.~Zhao, N.~Wang, and J.~Chen, ``Using squeeze-and-excitation vision transformer with local feature fusion for ship classification in \mbox{SAR} images,'' in \emph{IGARSS 2023 - 2023 IEEE International Geoscience and Remote Sensing Symposium}, 2023, pp. 7499--7502.

\bibitem{10189816}
H.~Zheng, Z.~Hu, L.~Yang, A.~Xu, M.~Zheng, C.~Zhang, and K.~Li, ``Multifeature collaborative fusion network with deep supervision for \mbox{SAR} ship classification,'' \emph{IEEE Transactions on Geoscience and Remote Sensing}, vol.~61, pp. 1--14, 2023.

\bibitem{10239480}
Y.~Liu, F.~Zhang, L.~Ma, and F.~Ma, ``Long-tailed \mbox{SAR} target recognition based on expert network and intraclass resampling,'' \emph{IEEE Geoscience and Remote Sensing Letters}, vol.~20, pp. 1--5, 2023.

\bibitem{10242360}
Y.~Guan, X.~Zhang, S.~Chen, G.~Liu, Y.~Jia, Y.~Zhang, G.~Gao, J.~Zhang, Z.~Li, and C.~Cao, ``Fishing vessel classification in \mbox{SAR} images using a novel deep learning model,'' \emph{IEEE Transactions on Geoscience and Remote Sensing}, vol.~61, pp. 1--21, 2023.

\bibitem{10246308}
G.~Gao, Y.~Dai, X.~Zhang, D.~Duan, and F.~Guo, ``\mbox{ADCG}: A cross-modality domain transfer learning method for synthetic aperture radar in ship automatic target recognition,'' \emph{IEEE Transactions on Geoscience and Remote Sensing}, vol.~61, pp. 1--14, 2023.

\bibitem{10315181}
Y.~Shang, W.~Pu, C.~Wu, D.~Liao, X.~Xu, C.~Wang, Y.~Huang, Y.~Zhang, J.~Wu, J.~Yang, and J.~Wu, ``\mbox{HDSS-Net}: A novel hierarchically designed network with spherical space classifier for ship recognition in \mbox{SAR} images,'' \emph{IEEE Transactions on Geoscience and Remote Sensing}, vol.~61, pp. 1--20, 2023.

\bibitem{10497597}
H.~Zhang, Y.~Lv, J.~Zhang, Q.~Wu, and Z.~Wen, ``Rendering-inspired cross-source feature disentanglement for domain adaptation- based sar ship classification,'' \emph{IEEE Geoscience and Remote Sensing Letters}, vol.~21, pp. 1--5, 2024.

\bibitem{10457854}
N.~Xie, T.~Zhang, W.~Guo, Z.~Zhang, and W.~Yu, ``Dual branch deep network for ship classification of dual-polarized sar images,'' \emph{IEEE Transactions on Geoscience and Remote Sensing}, vol.~62, pp. 1--15, 2024.

\bibitem{10487903}
S.~Bhattacharjee, P.~Shanmugam, and S.~Das, ``Attention-guided convolution neural network assisted with handcrafted features for ship classification in low-resolution sentinel-1 sar image data,'' \emph{IEEE Access}, vol.~12, pp. 48\,668--48\,685, 2024.

\bibitem{10446741}
X.~Feng, H.~Zheng, Z.~Hu, L.~Yang, and M.~Zheng, ``Dual-stream contrastive predictive network with joint handcrafted feature view for \mbox{SAR} ship classification,'' in \emph{ICASSP 2024 - 2024 IEEE International Conference on Acoustics, Speech and Signal Processing (ICASSP)}, 2024, pp. 7810--7814.

\bibitem{rs16071299}
\BIBentryALTinterwordspacing
L.~Wang, Y.~Qi, P.~T. Mathiopoulos, C.~Zhao, and S.~Mazhar, ``An improved \mbox{SAR} ship classification method using text-to-image generation-based data augmentation and squeeze and excitation,'' \emph{Remote Sensing}, vol.~16, no.~7, 2024. [Online]. Available: \url{https://www.mdpi.com/2072-4292/16/7/1299}
\BIBentrySTDinterwordspacing

\bibitem{10547709}
Z.~Liu, K.~Li, L.~Wang, and Z.~Zhang, ``Multi-scale alignment domain adaptation for ship classification in multi-resolution \mbox{SAR} images,'' \emph{IEEE Transactions on Automation Science and Engineering}, vol.~22, pp. 4051--4062, 2025.

\bibitem{s24247954}
\BIBentryALTinterwordspacing
A.~Toumi, J.-C. Cexus, A.~Khenchaf, and M.~Abid, ``A combined \mbox{CNN-LSTM} network for ship classification on \mbox{SAR} images,'' \emph{Sensors}, vol.~24, no.~24, 2024. [Online]. Available: \url{https://www.mdpi.com/1424-8220/24/24/7954}
\BIBentrySTDinterwordspacing

\bibitem{10642068}
N.~Xie, M.~Xiong, F.~Wei, J.~Li, T.~Zhang, and W.~Yu, ``\mbox{MFSAF}: A plug-and-play module for \mbox{SAR} ship classification,'' in \emph{IGARSS 2024 - 2024 IEEE International Geoscience and Remote Sensing Symposium}, 2024, pp. 9075--9079.

\bibitem{10440349}
N.~Xie, T.~Zhang, F.~Wei, and W.~Yu, ``Pfdn: A polarimetric feature-guided deep network for dual-polarized \mbox{SAR} ship classification,'' \emph{IEEE Geoscience and Remote Sensing Letters}, vol.~21, pp. 1--5, 2024.

\bibitem{rs16183479}
\BIBentryALTinterwordspacing
J.~He, R.~Sun, Y.~Kong, W.~Chang, C.~Sun, G.~Chen, Y.~Li, Z.~Meng, and F.~Wang, ``\mbox{CPINet}: Towards a novel cross-polarimetric interaction network for dual-polarized \mbox{SAR} ship classification,'' \emph{Remote Sensing}, vol.~16, no.~18, 2024. [Online]. Available: \url{https://www.mdpi.com/2072-4292/16/18/3479}
\BIBentrySTDinterwordspacing

\bibitem{10671463}
D.~Li and Z.~Zhao, ``\mbox{SAR} ship classification based on second-order synchrosqueezing transform and \mbox{HOG} feature,'' in \emph{2024 9th International Conference on Signal and Image Processing (ICSIP)}, 2024, pp. 585--590.

\bibitem{10412205}
X.~Zhang, S.~Feng, C.~Zhao, Z.~Sun, S.~Zhang, and K.~Ji, ``\mbox{MGSFA-Net}: Multiscale global scattering feature association network for \mbox{SAR} ship target recognition,'' \emph{IEEE Journal of Selected Topics in Applied Earth Observations and Remote Sensing}, vol.~17, pp. 4611--4625, 2024.

\bibitem{10641086}
A.~A. Al~Hinai and R.~Guida, ``Uncertainty estimation in \mbox{Bayesian} convolutional neural networks for \mbox{SAR} ship classification,'' in \emph{IGARSS 2024 - 2024 IEEE International Geoscience and Remote Sensing Symposium}, 2024, pp. 7604--7608.

\bibitem{10578024}
Z.~Cui, L.~Mou, Z.~Zhou, K.~Tang, Z.~Yang, Z.~Cao, and J.~Yang, ``Feature joint learning for \mbox{SAR} target recognition,'' \emph{IEEE Transactions on Geoscience and Remote Sensing}, vol.~62, pp. 1--20, 2024.

\bibitem{10446152}
B.~Xu, H.~Zheng, Z.~Hu, L.~Yang, M.~Zheng, X.~Feng, and W.~Lin, ``Double reverse regularization network based on self-knowledge distillation for \mbox{SAR} object classification,'' in \emph{ICASSP 2024 - 2024 IEEE International Conference on Acoustics, Speech and Signal Processing (ICASSP)}, 2024, pp. 7800--7804.

\bibitem{10701968}
C.~Awais and M.~Reggiannini, ``Deep learning for \mbox{SAR} ship classification: Focus on unbalanced datasets and inter-dataset generalization,'' in \emph{2024 International Conference on Electromagnetics in Advanced Applications (ICEAA)}, 2024, pp. 1--1.

\bibitem{10908205}
S.~Zhao, W.~Li, F.~Shen, and M.~You, ``Ln-scnet: A lightweight convolutional neural network for \mbox{SAR} ship classification,'' \emph{IEEE Access}, vol.~13, pp. 39\,394--39\,404, 2025.

\bibitem{10890957}
G.~Gao, Y.~He, J.~Zhao, S.~Li, M.~Wang, G.~Yang, and X.~Zhang, ``\mbox{FDC-TA-DSN} ship classification model and dataset construction based on complex-valued \mbox{SAR},'' \emph{IEEE Journal of Selected Topics in Applied Earth Observations and Remote Sensing}, vol.~18, pp. 7034--7047, 2025.

\bibitem{11045289}
Z.~Han and H.~Lang, ``Unsupervised \mbox{SAR} fine-grained ship classification via spherical metric refinement with deep subdomain adaptation,'' \emph{IEEE Journal of Selected Topics in Applied Earth Observations and Remote Sensing}, vol.~18, pp. 16\,003--16\,019, 2025.

\bibitem{Awais_2025_WACV}
C.~M. Awais, M.~Reggiannini, and D.~Moroni, ``A framework for imbalanced \mbox{SAR} ship classification: Curriculum learning weighted loss functions and a novel evaluation metric,'' in \emph{Proceedings of the Winter Conference on Applications of Computer Vision (WACV) Workshops}, February 2025, pp. 1570--1578.

\bibitem{10937719}
J.~Zheng, J.~Cao, and X.~Hu, ``Msf-at: A study on ship \mbox{SAR} image classification based on multi-scale feature and attention mechanism,'' \emph{IEEE Access}, vol.~13, pp. 55\,467--55\,475, 2025.

\bibitem{11043629}
C.~M. Awais, M.~Reggiannini, and D.~Moroni, ``Image quality vs performance in super-resolution for \mbox{SAR} ship classification,'' in \emph{2025 IEEE International Symposium on Circuits and Systems (ISCAS)}, 2025, pp. 1--5.

\bibitem{11018437}
W.-L. Tseng, Y.-L. Wu, M.-C. Lee, and H.-H. Shuai, ``\mbox{RSMAE}: Radiometric resolution and scale-aware masked autoencoder for \mbox{SAR} ship recognition,'' \emph{IEEE Geoscience and Remote Sensing Letters}, vol.~22, pp. 1--5, 2025.

\bibitem{10960691}
N.~Xie, T.~Zhang, L.~Zhang, J.~Chen, F.~Wei, and W.~Yu, ``\mbox{VLF-SAR}: A novel vision-language framework for few-shot \mbox{SAR} target recognition,'' \emph{IEEE Transactions on Circuits and Systems for Video Technology}, pp. 1--1, 2025.

\bibitem{awais2025classification}
C.~M. Awais, M.~Reggiannini, D.~Moroni, and O.~Karakus, ``A classification-aware super-resolution framework for ship targets in \mbox{SAR} imagery,'' \emph{arXiv preprint arXiv:2508.06407}, 2025.

\bibitem{awais2025feature}
------, ``Feature-space oversampling for addressing class imbalance in \mbox{SAR} ship classification,'' \emph{arXiv preprint arXiv:2508.06420}, 2025.

\bibitem{10254064}
E.~Nehary, A.~Dey, S.~Rajan, B.~Balaji, A.~Damini, and R.~Chanchlani, ``Synthetic aperture radar-based ship classification using \mbox{CNN} and traditional handcrafted features,'' in \emph{2023 IEEE Sensors Applications Symposium (SAS)}, 2023, pp. 01--06.

\bibitem{EMMENS2021114975}
\BIBentryALTinterwordspacing
T.~Emmens, C.~Amrit, A.~Abdi, and M.~Ghosh, ``The promises and perils of automatic identification system data,'' \emph{Expert Systems with Applications}, vol. 178, p. 114975, 2021. [Online]. Available: \url{https://www.sciencedirect.com/science/article/pii/S0957417421004164}
\BIBentrySTDinterwordspacing

\bibitem{li2017opensarship}
B.~Li, B.~Liu, L.~Huang, W.~Guo, Z.~Zhang, and W.~Yu, ``\mbox{OpenSARShip} 2.0: A large-volume dataset for deeper interpretation of ship targets in \mbox{Sentinel-1} imagery,'' in \emph{2017 SAR in Big Data Era: Models, Methods and Applications (BIGSARDATA)}.\hskip 1em plus 0.5em minus 0.4em\relax IEEE, 2017, pp. 1--5.

\bibitem{9884456}
W.~Yang, Z.~Ma, and Y.~Shi, ``\mbox{SAR} image super-resolution based on artificial intelligence,'' in \emph{IGARSS 2022 - 2022 IEEE International Geoscience and Remote Sensing Symposium}, 2022, pp. 4643--4646.

\bibitem{rs16010018}
\BIBentryALTinterwordspacing
L.~Bu, J.~Zhang, Z.~Zhang, Y.~Yang, and M.~Deng, ``Deep learning for integrated speckle reduction and super-resolution in multi-temporal \mbox{SAR},'' \emph{Remote Sensing}, vol.~16, no.~1, 2024. [Online]. Available: \url{https://www.mdpi.com/2072-4292/16/1/18}
\BIBentrySTDinterwordspacing

\bibitem{Smith_2022}
\BIBentryALTinterwordspacing
J.~W. Smith, Y.~Alimam, G.~Vedula, and M.~Torlak, ``A vision transformer approach for efficient near-field \mbox{SAR} super-resolution under array perturbation,'' in \emph{2022 IEEE Texas Symposium on Wireless and Microwave Circuits and Systems (WMCS)}.\hskip 1em plus 0.5em minus 0.4em\relax IEEE, Apr. 2022. [Online]. Available: \url{http://dx.doi.org/10.1109/WMCS55582.2022.9866326}
\BIBentrySTDinterwordspacing

\bibitem{hoffer2015deep}
E.~Hoffer and N.~Ailon, ``Deep metric learning using triplet network,'' in \emph{Similarity-based pattern recognition: third international workshop, SIMBAD 2015, Copenhagen, Denmark, October 12-14, 2015. Proceedings 3}.\hskip 1em plus 0.5em minus 0.4em\relax Springer, 2015, pp. 84--92.

\bibitem{afMSTAROverview}
``The \mbox{Air Force} moving and stationary target recognition database. available online:,'' \url{https://www.sdms.afrl.af.mil/index.php?collection=mstar}, [Accessed 26-06-2024].

\bibitem{wen2016discriminative}
Y.~Wen, K.~Zhang, Z.~Li, and Y.~Qiao, ``A discriminative feature learning approach for deep face recognition,'' in \emph{Computer vision--ECCV 2016: 14th European conference, amsterdam, the netherlands, October 11--14, 2016, proceedings, part VII 14}.\hskip 1em plus 0.5em minus 0.4em\relax Springer, 2016, pp. 499--515.

\bibitem{9784428}
H.~Lang, G.~Yang, C.~Li, and J.~Xu, ``Multisource heterogeneous transfer learning via feature augmentation for ship classification in \mbox{SAR} imagery,'' \emph{IEEE Transactions on Geoscience and Remote Sensing}, vol.~60, pp. 1--14, 2022.

\bibitem{krizhevsky2012imagenet}
A.~Krizhevsky, I.~Sutskever, and G.~E. Hinton, ``Imagenet classification with deep convolutional neural networks,'' \emph{Advances in neural information processing systems}, vol.~25, 2012.

\bibitem{simonyan2014very}
K.~Simonyan and A.~Zisserman, ``Very deep convolutional networks for large-scale image recognition,'' \emph{arXiv preprint arXiv:1409.1556}, 2014.

\bibitem{he2016deep}
K.~He, X.~Zhang, S.~Ren, and J.~Sun, ``Deep residual learning for image recognition,'' in \emph{Proceedings of the IEEE conference on computer vision and pattern recognition}, 2016, pp. 770--778.

\bibitem{huang2017densely}
G.~Huang, Z.~Liu, L.~Van Der~Maaten, and K.~Q. Weinberger, ``Densely connected convolutional networks,'' in \emph{Proceedings of the IEEE conference on computer vision and pattern recognition}, 2017, pp. 4700--4708.

\bibitem{szegedy2015going}
C.~Szegedy, W.~Liu, Y.~Jia, P.~Sermanet, S.~Reed, D.~Anguelov, D.~Erhan, V.~Vanhoucke, and A.~Rabinovich, ``Going deeper with convolutions,'' in \emph{Proceedings of the IEEE conference on computer vision and pattern recognition}, 2015, pp. 1--9.

\bibitem{jiang2021transgan}
Y.~Jiang, S.~Chang, and Z.~Wang, ``Transgan: Two pure transformers can make one strong gan, and that can scale up,'' \emph{Advances in Neural Information Processing Systems}, vol.~34, pp. 14\,745--14\,758, 2021.

\bibitem{xiao2025foundation}
A.~Xiao, W.~Xuan, J.~Wang, J.~Huang, D.~Tao, S.~Lu, and N.~Yokoya, ``Foundation models for remote sensing and earth observation: A survey,'' \emph{IEEE Geoscience and Remote Sensing Magazine}, 2025.

\bibitem{selvaraju2017grad}
R.~R. Selvaraju, M.~Cogswell, A.~Das, R.~Vedantam, D.~Parikh, and D.~Batra, ``Grad-cam: Visual explanations from deep networks via gradient-based localization,'' in \emph{Proceedings of the IEEE international conference on computer vision}, 2017, pp. 618--626.

\bibitem{lundberg2017unified}
S.~M. Lundberg and S.-I. Lee, ``A unified approach to interpreting model predictions,'' \emph{Advances in neural information processing systems}, vol.~30, 2017.

\bibitem{ribeiro2016should}
M.~T. Ribeiro, S.~Singh, and C.~Guestrin, ```\mbox{Why} should i trust you?' \mbox{Explaining} the predictions of any classifier,'' in \emph{Proceedings of the 22nd ACM SIGKDD international conference on knowledge discovery and data mining}, 2016, pp. 1135--1144.

\end{thebibliography}
\bibliographystyle{IEEEtran} 

\newpage

\section{Biography Section}
\vspace{-10pt}
\begin{IEEEbiography}[{\includegraphics[width=1in,height=1.15in,clip,keepaspectratio]{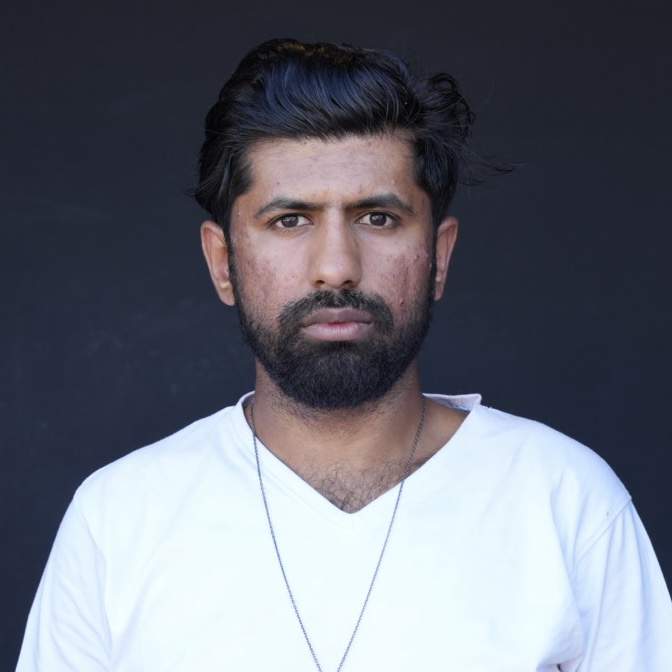}}]{Ch Muhammad Awais}
is a Ph.D. student at the University of Pisa, specializing in SAR ship classification. Hailing from Pakistan, he has been engaged in DL research for the past six years and has worked on SAR ship classification for the last two years.
\end{IEEEbiography}
\vspace{-40pt}
\begin{IEEEbiography}[{\includegraphics[width=1in,height=1.15in,clip,keepaspectratio]{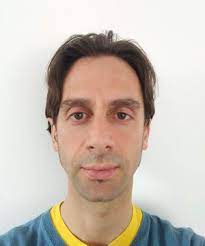}}]{Marco Reggiannini} received his M.Sc. in Physics and a Ph.D. in Automation Engineering from the University of Pisa. Currently, he is a researcher at the Signals and Imaging Lab at ISTI-CNR, with more than ten years of experience devoted to multisensor imagery analysis.
\end{IEEEbiography}
\vspace{-40pt}
\begin{IEEEbiography}[{\includegraphics[width=1in,height=1.15in,clip,keepaspectratio]{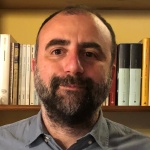}}]{ Davide Moroni} received his M.Sc. in Mathematics (Hons.) from the University of Pisa and a Ph.D. in Mathematics from the University of Rome La Sapienza. He is a Senior Researcher and Head of the Signals and Images Lab at ISTI-CNR, specializing in image processing and computer vision.
\end{IEEEbiography}
\vspace{-40pt}
\begin{IEEEbiography}[{\includegraphics[width=1in,height=1.15in,clip,keepaspectratio]{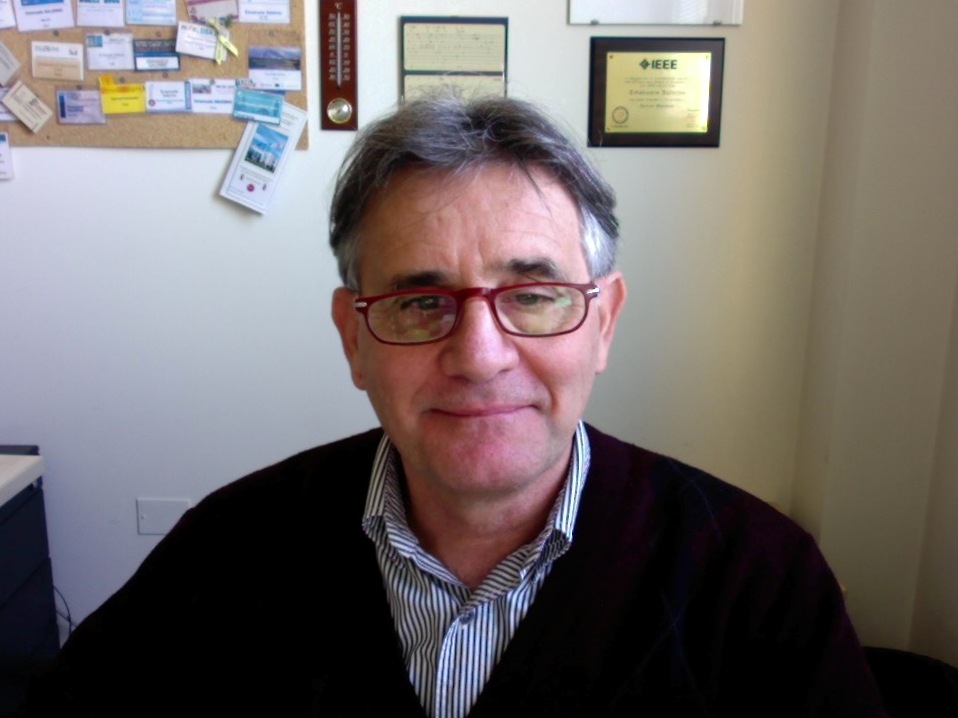}}]{Emanuele Salerno} graduated in Electronic Engineering from the University of Pisa in 1985, joined CNR in 1987 and is now a Senior Research Associate at ISTI-CNR. His expertise spans SAR image processing, cultural heritage technologies, and computational biology, with a significant academic and research background in signal and image processing.
\end{IEEEbiography}

\vfill

\end{document}